\documentclass[10pt,journal,compsoc]{assets/IEEEtran}

\ifCLASSOPTIONcompsoc
  \usepackage[nocompress]{cite}
\else
  \usepackage{cite}
\fi

\ifCLASSINFOpdf
\usepackage{tikz}
\graphicspath{figures}
\DeclareGraphicsExtensions{.pdf,.png}
\else
\fi

\usepackage{array}

\usepackage{booktabs}
\usepackage{tabularx}
\usepackage{multirow}
\usepackage[flushleft]{threeparttable}
\newcolumntype{C}{>{\Centering\arraybackslash}X} %

\usepackage{url}

\newcommand{\nMLTotal}{950}
\newcommand{\nDLTotal}{233}
\newcommand{\avgRelIncr}{25\%}
\newcommand{\avgRelIncrDL}{119\%}
\newcommand{\prcDeepTotal}{52\%}

\newcommand{\nFER}{158}
\newcommand{\nBody}{4}
\newcommand{\nSER}{82}
\newcommand{\nEEG}{18}
\newcommand{\nPeri}{7}
\newcommand{\nOther}{3}
\newcommand{\nFERSpatial}{141}
\newcommand{\nFERTemporal}{36}
\newcommand{\nFERJoint}{19}
\newcommand{\nBodySpatial}{-}
\newcommand{\nBodyTemporal}{4}
\newcommand{\nBodyJoint}{2}
\newcommand{\nSERSpatial}{35}
\newcommand{\nSERTemporal}{64}
\newcommand{\nSERJoint}{17}
\newcommand{\nEEGSpatial}{4}
\newcommand{\nEEGTemporal}{17}
\newcommand{\nEEGJoint}{3}
\newcommand{\nPeriSpatial}{-}
\newcommand{\nPeriTemporal}{6}
\newcommand{\nPeriJoint}{4}
\newcommand{\nOtherSpatial}{-}
\newcommand{\nOtherTemporal}{2}
\newcommand{\nOtherJoint}{3}
\newcommand{\nPhysioTiii}{four}
\newcommand{\nSpatial}{173}
\newcommand{\nTemporal}{105}
\newcommand{\nJoint}{21}

\newcommand{\prcFERSpatialRegionI}{56\%}

\newcommand{\nAudioVisual}{18}
\newcommand{\nAVP}{three}

\newcommand{\nVGG}{23}
\newcommand{\nAlexNet}{18}
\newcommand{\nCNNLSTM}{21}
\newcommand{\nSERCNNLSTM}{eight}

\newcommand{\nComparison}{150}
\newcommand{\prcDeepBetter}{95\%}

\newcommand{\nSpatialComparison}{103}
\newcommand{\prcSpatialDeepBetter}{93\%}
\newcommand{\nTemporalComparison}{73}
\newcommand{\prcTemporalDeepBetter}{92\%}
\newcommand{\nJointComparison}{16}
\newcommand{\prcJointDeepBetter}{69\%}

\newcommand{\prcFERSpatialDeepBetter}{93\%}
\newcommand{\prcSERSpatialDeepBetter}{90\%}
\newcommand{\prcFERTemporalDeepBetter}{94\%}
\newcommand{\prcSERTemporalDeepBetter}{91\%}

\newcommand{\prcFERSpatialDropoutFC}{57\%}
\newcommand{\prcFERSpatialDropoutConv}{12\%}

\newcommand{\prcSERSpatialDropoutFC}{43\%}
\newcommand{\prcSERSpatialDropoutConv}{14\%}

\newcommand{\prcPrivateDB}{11\%}
\newcommand{\nDBTotal}{77}

\newcommand{\nCatDatabase}{58}
\newcommand{\nDimDatabase}{14}
\newcommand{\nBothDatabase}{4}
\newcommand{\prcCatDatabase}{75\%}
\newcommand{\prcDimDatabase}{18\%}
\newcommand{\prcBothDatabase}{5.2\%}

\newcommand{\nCatStudies}{190}
\newcommand{\nDimStudies}{37}
\newcommand{\nBothStudies}{6}
\newcommand{\prcCatStudies}{82\%}
\newcommand{\prcDimStudies}{16\%}
\newcommand{\prcBothStudies}{2.6\%}

\newcommand{\prcCatSpatial}{92\%}

\newcommand{\prcDimSpatial}{10\%}
\newcommand{\prcDimTemporal}{30\%}
\newcommand{\prcDimJoint}{52\%}

\newcommand{\ILSVRC}{ILSVRC}

\begin{document}

\title{Deep Learning for Human Affect Recognition:\\ Insights and New Developments}

\author{Philipp~V.~Rouast,~\IEEEmembership{Student Member,~IEEE,}
        Marc~T.~P.~Adam,
        and~Raymond~Chiong,~\IEEEmembership{Senior Member,~IEEE}
\IEEEcompsocitemizethanks{\IEEEcompsocthanksitem P. V. Rouast, M. T. P. Adam and R. Chiong are with the School of Electrical Engineering and Computing, University of Newcastle, Callaghan, NSW 2308, Australia.\protect\\
E-mail: philipp.rouast@uon.edu.au}%
\thanks{Manuscript submitted Feb. 26, 2018; revised Aug. 31, 2018, Nov. 16, 2018.}}

\IEEEpubid{ \begin{minipage}{\textwidth} \tiny
This article has been accepted for publication in a future issue of this journal, but has not been fully edited. Content may change prior to final publication. Citation information: DOI 10.1109/TAFFC.2018.2890471, IEEE \\
1949-3045 \copyright~2018 IEEE. Personal use is permitted, but republication/redistribution requires IEEE permission. See \url{http://www.ieee.org/publications_standards/publications/rights/index.html} for more information.
\end{minipage}}

\markboth{Journal of \LaTeX\ Class Files,~Vol.~14, No.~8, August~2015}%
{Rouast \MakeLowercase{\textit{et al.}}: Deep Learning for Human Affect Recognition: Insights and New Developments}

\IEEEtitleabstractindextext{%
\begin{abstract}
Automatic human affect recognition is a key step towards more natural human-computer interaction.
Recent trends include recognition in the wild using a fusion of audiovisual and physiological sensors, a challenging setting for conventional machine learning algorithms.
Since 2010, novel deep learning algorithms have been applied increasingly in this field.
In this paper, we review the literature on human affect recognition between 2010 and 2017, with a special focus on approaches using deep neural networks.
By classifying a total of {\nMLTotal} studies according to their usage of shallow or deep architectures, we are able to show a trend towards deep learning.
Reviewing a subset of {\nDLTotal} studies that employ deep neural networks, we comprehensively quantify their applications in this field.
We find that deep learning is used for learning of (i) spatial feature representations, (ii) temporal feature representations, and (iii) joint feature representations for multimodal sensor data.
Exemplary state-of-the-art architectures illustrate the progress.
Our findings show the role deep architectures will play in human affect recognition, and can serve as a reference point for researchers working on related applications.
\end{abstract}

}

\maketitle

\IEEEdisplaynontitleabstractindextext

%
\IEEEpeerreviewmaketitle

\IEEEraisesectionheading{\section{Introduction}}\label{sec:introduction}

\IEEEPARstart{H}{uman-computer interaction} (HCI) is seeing a gradual shift from the computer-centered approach of the past to a more user-centered approach.
Commercial success has made user-friendly input methods, portable devices and multi-sensor availability a new standard in personal computing.
Despite the progress, it has been argued that HCI is still lacking a central element of human-human interaction: The communication of information through affective display \cite{picard1997affective}.
The term \textit{affective computing} was coined by Rosalind Picard in 1995 \cite{picard1995affective}, inspired by findings from neuroscience, psychology, and cognitive science highlighting the important role affect plays in intelligent behavior.
It encompasses efforts to (i) automatically recognize human affect, and (ii) generate corresponding responses by the computer, providing a richer context for HCI outcomes.
Applications range from education \cite{dmello2007toward} and health care \cite{lisetti2003developing} to entertainment \cite{yannakakis2011experience} and embodied agents \cite{de2003greta}.

In human interactions, a significant amount of information is not communicated explicitly, but through the way we speak, our facial expressions, gestures, and other means.
Initial research on affect recognition\footnote{Unless stated otherwise, in this article we use the term affect recognition to refer to human affect recognition.} focused mainly on unimodal approaches, with facial expression recognition (FER) and speech emotion recognition (SER) gaining most attention and highest accuracies \cite{pantic2003toward}.
Psychophysiological measures were also shown to contain information about affective states \cite{picard2001toward}.
Public database availability has improved since the 2000s \cite{calvo2010affect} and multimodal sensor combination was found to improve recognition accuracy and robustness \cite{jaimes2007multimodal}.
As good performance was achieved on posed databases, the focus started to shift towards more realistic, spontaneous displays of affective behavior \cite{zeng2009survey}.
These trends are exemplified in the annual competitions \textit{Emotion Recognition in the Wild} (EmotiW) \cite{dhall2016emotiw} and \textit{Audio Video Emotion Challenge} (AVEC) \cite{valstar2016avec}.
Since 2010, \textit{deep learning} methods have been applied to affect recognition problems across multiple modalities and led to improvements in accuracy, including winning performances at EmotiW \cite{fan2016video}, \cite{li2016happiness}, \cite{kim2015hierarchical} and AVEC \cite{chen2017multimodal}, \cite{brady2016multi}, \cite{he2015multimodal}.

In this paper, we review the state of research on affect recognition using deep learning.
We start by giving an introduction to deep learning, the reasoning behind its application in artificial intelligence (AI), and to the most relevant architectures.
We then break down the challenges faced in affect recognition research, and outline why models incorporating deep learning are useful in meeting these.
Finally, we go on to discuss the directions research has taken since 2010 and in what way deep learning methods are utilized.
Our contributions to the field are:

\begin{itemize}
	\item By conducting a comprehensive literature search, and classification of a total of {\nMLTotal} studies, we are able to identify and measure a trend towards use of deep neural networks for affect recognition;
	\item reviewing {\nDLTotal} studies that employ deep learning, we identify the main application areas as being the learning of (i) \textit{spatial feature representations}, (ii) \textit{temporal feature representations}, and (iii) \textit{joint feature representations} for multimodal sensor data;  
	\item we discuss and give exemplary illustrations of how deep neural networks are applied on visual, auditory, and physiological sensor data;
	\item we provide an overview of the most relevant databases by quantifying their use across {\nDLTotal} studies, including large databases established since 2016; and
	\item we discuss open issues and research directions.
\end{itemize}

\section{Deep learning for affect recognition}\label{sec:deep-learning-for-affect-recognition}

Automatic affect recognition relies on visual and auditory perception \cite{bengio2007scaling}, which are examples of abilities commonly expected of AI.
A class of methods known as \textit{machine learning} has turned out to be effective in handling the desired perception tasks.
By enabling computers to learn directly from examples, these algorithms overcome the need to provide an explicit model.

Although perception tasks such as visual and auditory affect recognition seem intuitive to humans, some characteristics rooted in their real-world origin make them hard problems to solve:
They require understanding of problems characterized by highly varying functions in terms of the input;
another common characteristic is the high dimensionality of examples in the form of images and audio files.

In this context, a common challenge for traditional machine learning approaches is the \textit{curse of dimensionality}, a phenomenon where the higher the number of dimensions used to represent the data, the less effective conventional computational and statistical methods become \cite{bengio2005curse}.
With a very high number of dimensions, it becomes increasingly difficult to comprehensively sample all possible combinations, resulting in vast unexplored regions in the feature space.
To circumvent this problem, a straightforward and widely used solution is to project the high-dimensional data into a lower-dimensional space through approaches such as feature selection.
Machine learning algorithms with so-called \textit{shallow} architectures, such as kernel methods and single-layer neural networks, can then be efficiently applied for modeling purposes.
However, when considering computational and statistical efficiency as well as human involvement, it has been suggested that shallow architectures may not be the most efficient way to approach challenging learning problems such as affect recognition \cite{bengio2007scaling}.
Hence, in 2010 researchers have started to explore the application of \textit{deep} architectures for affect recognition.

\subsection{Deep learning}\label{sec:deep-learning}

To distinguish between \textit{shallow} and \textit{deep} machine learning models, one can think of their architectures as subsequent layers of hierarchical computation.
We will use the notion of architecture \textit{depth} to denote the number of computational layers in an architecture\footnote{This excludes the input layer, which lacks learnable parameters.}. 
Traditional learning algorithms can generally be represented as two layers of computation, where the first layer consists of template matchers or simple trainable basis functions, and the second layer is a weighted sum \cite{bengio2007scaling}.
This is why we talk of them as \textit{shallow} architectures, when comparing them with \textit{deep} neural networks, which consist of three or more layers.
Although there is no universally agreed upon rule to determine depth or distinguish between shallow and deep architectures \cite[ch.1]{goodfellow2016deep}, a cutoff of three or more layers is commonly used \cite{hinton2006fast}, \cite{lecun2015deep}, \cite{schmidhuber2015deep}.

\subsubsection{Deep neural networks}\label{sec:dnn}

The design of deep neural networks (DNNs) is loosely inspired by biological neural networks.
A typical example is the deep feedforward network (or multi-layer perceptron).
It consists of multiple layers of processing units (``neurons''): An input layer, multiple hidden layers, and an output layer.
Units in adjacent layers can have weighted connections.
We speak of \textit{fully-connected} DNNs if there are connections between all pairs of units in adjacent layers.
Information in the network flows forward through these connections, each unit computing its \textit{activation} as a function of its inputs.
Units in hidden layers introduce a nonlinearity in the process.
By adjusting the connection weights, a DNN can effectively learn a \textit{feature representation} of its input data in each layer's unit activations.
A well-trained DNN learns a deep hierarchy of \textit{distributed representations}\footnote{This implies a many-to-many relationship between learned concepts and units representing them.}.
This enables the network to learn very expressive representations capturing a large number of possible input configurations \cite{bengio2009learning}.

Some of the key advantages of deep architectures are derived from their depth.
Increasing depth promotes re-use of learned features \cite{bengio2013representation}.
On a related note, the deep hierarchy of feature representations allows learning at different levels of abstraction building on top of each other.
Here, higher levels of abstraction are generally associated with invariance to local changes of the input \cite{bengio2013representation}.

While the theoretical advantages of such deep architectures were known for some time, progress was held back by the difficulty of training them \cite{schmidhuber2015deep}.
An initial breakthrough in 2006 \cite{hinton2006fast} showed that the so-called \textit{vanishing gradient problem} in training DNNs can be overcome by unsupervised pre-training.
Multiple strategies of alleviating the problem are known today, including (i) architectures unaffected by it \cite{hochreiter1997long}, \cite{he2016deep}, (ii) improved optimizers \cite{martens2011learning}, (iii) certain training and design choices \cite{glorot2011deep}, \cite{sutskever2013importance}, \cite{ioffe2015batch}, and (iv) use of powerful computing systems, especially GPU-based.
Three interrelated factors drive the continued success of DNNs:

\begin{itemize}
	\item \textit{Increased learning capacity\footnote{Due to the complexity of deep learning algorithms, it is difficult to explicitly determine their learning capacity \cite[ch.5.2]{goodfellow2016deep}, but it can be thought of as being related to the number of parameters and layers \cite{sun2017revisiting}.}.}
		A central theme in the evolution of deep models has been a link between better generalization ability and increased number of parameters, especially when growing models \textit{deeper} rather than \textit{wider} \cite[ch.6.4]{goodfellow2016deep}.
		This applies as long as training is feasible and the dataset is large enough to take advantage of the architecture \cite{he2016deep}.
	\item \textit{Growing computing power.}
		Training of state-of-the-art deep learning models is an intense task involving millions of parameters to optimize.
		GPUs are particularly well suited for the operations involved, and specialized software is available.
		See the supplemental material (Section S4) for further information.
	\item \textit{Large datasets.}
		Part of the promise of deep learning is to take advantage of large datasets with millions of examples.
		An investigation into the effect of dataset size suggests a logarithmic relationship between performance on vision tasks and training data size \cite{sun2017revisiting}.
\end{itemize}

When dealing with data that are known to have a certain structure (e.g., spatial structure, temporal structure), DNNs can be modified to create more specialized architectures that take advantage of said structures.
In the following, we provide a brief overview of such specialized DNN architectures.

\subsubsection{Learning spatially: Convolutional neural networks}\label{sec:cnn}

Sensor recordings of natural scenes often inherently contain a spatial structure along some dimensions.
Convolutional neural networks (CNNs) introduce layers with specialized operations into DNNs, which take into account the spatial structure of the data to make the network more efficient.

The convolutional layer consists of several learnable kernels, which are convolved with the layer input to produce activations.
In terms of the layer's units, this can be interpreted as \textit{parameter sharing} between units, since the same kernel is applied over different spatial locations.
The kernel's receptive field is typically much smaller than the input, which leads to \textit{sparse connectivity} between units of adjacent layers \cite[ch.9.2]{goodfellow2016deep}.
In contrast to the fully-connected equivalent, this means that convolutional layers only establish connections between units that are spatially close, which dramatically reduces the number of parameters.
Since CNNs are inspired by the mammalian visual cortex \cite{hubel1962receptive}, we primarily see 2D convolutions applied to image data.
However, it is possible to apply convolutions along any dimension of the input data, including 1D (e.g., audio data) and 3D convolutions (e.g., video data).

After the convolution operation, a nonlinearity is applied, typically the rectified linear unit \cite{glorot2011deep}.
Pooling layers between convolutional layers facilitate nonlinear downsampling of layer activations, and make the network invariant to translations in the input \cite[ch.9.3]{goodfellow2016deep}.
Invariances to other transformations such as rotation or scaling are not directly implied by the architecture, but can be learned in convolutional layers. 
Finally, it is common to add one or two fully-connected layers after several alternating convolutional and pooling layers (e.g., \cite{krizhevsky2012imagenet}).

Compared to fully-connected DNNs, training CNNs is less difficult due to the reduced number of parameters.
They were trained successfully in the 1990s \cite{lecun1989handwritten}, while fully-connected DNNs were still believed to be too difficult to train \cite{schmidhuber2015deep}.
Since a record-breaking performance \cite{krizhevsky2012imagenet} at the \textit{ImageNet} Large Scale Visual Recognition Challenge (ILSVRC) \cite{russakovsky2015imagenet} in 2012, CNNs have attracted much attention from researchers and the mainstream media.

\subsubsection{Learning from sequences: Recurrent neural networks}\label{sec:rnn}

When learning from sequential data (e.g., audio, video), the goal is to capture temporal dynamics in an efficient way that allows generalization to sequences of arbitrary length.
This can be accomplished by sharing parameters across time, instead of re-learning them for every step.
As mentioned previously, CNNs can accomplish parameter sharing in a shallow way, by applying the same kernel at different points in time.
Recurrent neural networks (RNNs), on the other hand, introduce recurrent connections in time, allowing parameters to be shared in a deeper way \cite[ch.10]{goodfellow2016deep}.

A basic RNN extends the feedforward architecture by allowing recurrent connections to exist within layers.
Simply put, the previous model state can be regarded as an additional input at each temporal step, which allows the RNN to form a memory in its hidden state over information from all previous inputs \cite{ming2017understanding}.
RNNs have a representational advantage over hidden Markov models (HMMs), whose discrete hidden states limit their memory \cite{lipton2015critical}.
Even RNNs with a single hidden layer can be considered as very deep networks, which becomes clear when we imagine ``unrolling'' them along the time dimension.
It turns out that this depth makes training RNNs considerably more difficult, as gradients tend to vanish or explode during training---especially when processing long sequences.

To deal with this problem, multiple specialized RNN architectures with gated units have been proposed. 
\textit{Long short-term memory} (LSTM) RNNs \cite{hochreiter1997long} are successful at learning long-term dependencies by providing gate mechanisms to add and forget information selectively.
\textit{Gated recurrent units} (GRUs) are a gating mechanism proposed more recently in the context of sequence-to-sequence processing \cite{cho2014learning}.
RNNs, especially LSTMs, have had a profound impact on how sequences of data are processed \cite{lecun2015deep}.
They are incorporated in state-of-the-art AI systems, for example in automatic speech recognition (ASR) \cite{han2018densely}.

\subsubsection{Unsupervised learning models}\label{sec:unsupervised}

To learn useful feature representations from data, the most common approach today is \textit{supervised} learning:
Researchers provide the learning algorithm with clues on how to improve parameters, typically in the form of corresponding data labels and a loss function measuring how ``bad'' a representation is.
Taking a classification task for example, a useful representation would be one that makes the classes of interest linearly separable.
However, labeling data is expensive and large unlabeled datasets are easier to come by.
\textit{Unsupervised} learning algorithms attempt to learn useful representations without being given explicit clues such as data labels---instead, a form of regularization is introduced.

An autoencoder (AE) \cite{baldi1989neural} is a neural network that learns two functions, $f: x \mapsto z$ and $g: z \mapsto x$, to restore its input $x$ from an intermediate representation $z$.
The idea of the AE is to avoid identity mapping between $f$ and $g$, either by forcing $z$ to be of lower dimensionality than $x$ in basic AEs, or by other forms of regularization, such as restoring $x$ from a corrupted version $\tilde{x}$ of itself in denoising AEs.
When multiple layers are between input or output and intermediate representation in an AE, we speak of stacked autoencoders (SAEs).
Restricted Boltzmann machines (RBMs) \cite{smolensky1986information} are undirected probabilistic models that learn a representation of their input in a layer of latent units.
Deep Boltzmann machines (DBMs) \cite{salakhutdinov2010efficient} consist of several layers of latent units with undirected connections. 
A deep belief network (DBN) \cite{hinton2006fast} on the other hand consists of several layers of directed connections, with an RBM as its final layer.

In practical applications, the lines between supervised and unsupervised learning are often blurred \cite[p.105]{goodfellow2016deep}.
Unsupervised pre-training is a technique whereby single-layer unsupervised models, such as RBMs, are iteratively trained and stacked into a deep model \cite{hinton2006fast}.
Both supervised and unsupervised models can benefit from the learned hierarchy of feature representations:
By adding a classification layer and supervised fine-tuning, it was found that difficulties in training fully-connected DNNs could be overcome \cite{bengio2007greedy}.
This technique was responsible for the resurgence of deep learning since 2006, but has later gone out of fashion as it is no longer required for training fully-connected DNNs.
Another use is found in initializing deep unsupervised models, such as DBNs and DBMs.

\subsection{The notion of affect in affect recognition}\label{sec:affect}

The notions of affect and emotion\footnote{The terms \textit{affect} and \textit{emotion} are widely used synonymously in the context of affective computing \cite{pantic2005affective}.} are subjective in nature.
As a result, the question of how affective states should be represented is still an unsolved issue with no consensus reached in the literature.
Two differing views are dominant: The \textit{categorical} view of affect as discrete states, and the \textit{dimensional} view of affect where states are represented in a continuous-valued space.
Both categorical and dimensional models are also used to decode the nature of human emotion in specific brain regions (e.g., amygdala, insula), which is subject to an ongoing debate in affective neuroscience (see \cite{kragel2016decoding_b} for a review).
Affective computing has largely stayed agnostic on the debate regarding the appropriateness of these views \cite{calvo2010affect}.
In practice, the type of affect labels available in databases is often chosen to most naturally fit the available sensor data---\textit{categorical} if images or sequences are to be matched with a single affective state, and \textit{dimensional} for continuous affect prediction (see Section \ref{sec:databases}).

The notion of categorical affective states originated in Charles Darwin's research on the evolution of affect, and remains much discussed in the literature \cite{ortony1990basic}.
Most categorical models assume the existence of \textit{basic} affective states as building blocks of more complex ones \cite{ortony1990basic}.
In practice, Ekman's basic affective states derived from facial expressions \cite{ekman1969repertoire} are most commonly used in the context of affect recognition;
they comprise \textit{anger}, \textit{happiness}, \textit{surprise}, \textit{disgust}, \textit{sadness}, and \textit{fear}.
Others include Plutchik's wheel of emotions \cite{plutchik1980general}, and the basic model \cite{james1890principles}.
In application-driven efforts, \textit{non-basic} affective states are more commonly used, where specific user states and the intensity thereof are of interest---such as \textit{boredom} or \textit{frustration} in HCI \cite{calvo2010affect}, \cite{zeng2009survey}.
This approach sidesteps issues of theoretical validity to instead focus directly on the relevant non-basic affective states.

Dimensional models aim to avoid the restrictiveness of discrete states, and allow more flexible definition of affective states as points in a multi-dimensional space spanned by concepts such as affect intensity and positivity.
This addresses the notion that discrete categories may not fully reflect the complexity of affective states and the associated problems in labeling and evaluating affective displays (e.g., labeler agreement) \cite{zeng2009survey}, \cite{calvo2010affect}, \cite{gunes2013categorical}.
For affect recognition, the dimensional space is commonly operationalized as a regression task (e.g., for arousal \cite{trigeorgis2016adieu}, \cite{brady2016multi}) or as a classification task where the continuous space is discretized (e.g., for different levels of arousal, \cite{li2016emotion}, \cite{bugnon2017dimensional}).
Mappings can be established between dimensional and categorical models of affect (e.g., \cite{mollahosseini2017affectnet}).
The most commonly used example is Russell's circumplex model \cite{russell1980circumplex}, which consists of the two dimensions \textit{valence} and \textit{arousal}, sometimes extended by \textit{dominance} and \textit{likability}.

\subsection{Frontiers in affect recognition}\label{sec:challenges}

Research on affect recognition has seen considerable progress as the focus has shifted from the study of lab-based, acted databases to real-life scenarios since the 2000s \cite{zeng2009survey}.
In the 2010s, the focus has remained on such conditions, as research is exploring more complex models to better take advantage of available sensor data.
The annual competitions AVEC and EmotiW have been established to focus on multimodal affect recognition, encouraging researchers to benchmark model accuracy in a fair manner.
Multiple factors make building such models challenging:

\begin{itemize}
	\item \textit{Natural settings.}
	Data collected in uncontrolled real-life settings can introduce many additional sources of variation that need to be accounted for.
	Visually, these include occlusions, complex illumination conditions in different settings, spontaneous behavior and poses, as well as rigid movements.
	Audio may contain background noise, unclear speech, interruptions, and other artifacts.
	\item \textit{Temporal dynamics.}
	The temporal dimension is an integral element of affective display that has not yet been fully embraced by research on affect recognition, especially FER.
	Temporal dynamics can be rich with contextual information that could be captured by models, e.g., to distinguish between displays that are similar in the short term, and to assess the relative importance of specific segments \cite{pantic2003toward}.
	\item \textit{Multimodal sensor fusion.}
	Humans rely on multiple modalities when expressing and sensing affective states in social interactions.
	It seems natural that computers could benefit from the same variety of sensors \cite{pantic2005affective}.
	In fact, there has been an increased interest to design such multimodal systems \cite{dmello2015review}, and it is generally accepted that audiovisual sensor fusion can increase model robustness and accuracy.
	However, without knowledge of how exactly humans handle this problem, it is unclear how and at which level of abstraction the modalities need to be fused.
	\item \textit{Limited availability of labeled data.}
	The increase of factors that make learning difficult---such as model size and the nature of ambitious affect recognition tasks---has outpaced the increase in availability of labeled examples. 
	It is a challenging task to successfully train large DNNs with relatively few labeled data.
	Advanced regularization methods are necessary to avoid problems like overfitting.
	Transfer learning \cite{pan2010survey} attempts to transfer knowledge between emotion corpora and other domains such as object recognition.
	Furthermore, unsupervised and semi-supervised learning are being explored to access knowledge encoded in unlabeled datasets.
\end{itemize}

\subsection{Towards learning deep models of affect}\label{sec:towards}

While traditional human affect recognition systems follow standard machine learning approaches, it seems intuitive that these algorithms may benefit from being more reminiscent of how the human brain works.
Leading researchers in affective computing have supported this notion in the past, voicing the expectation that biologically inspired systems could be more suitable for affect recognition than human-engineered systems \cite{picard1997affective}, \cite{pantic2005affective}.

In light of the challenges discussed in Section \ref{sec:challenges}, current affect recognition tasks combine several characteristics of difficult perception problems in AI.
The variation in the sensor data is dominated by factors mostly unrelated to the task, (i) spatially (e.g., in image data and short-term audio segments), (ii) temporally (e.g., in image sequences and long-term audio), and (iii) distributed across modalities on different time scales.
In order to perform the desired classification and regression tasks in affect recognition, it is necessary to obtain useful feature representations from the available multimodal high-dimensional sensor data.
These feature representations should disentangle the underlying factors of variation in order to isolate the factors that distinguish affective displays \cite{rifai2012disentangling}.

Previous approaches rely on cleverly handcrafted features and shallow learning models to derive such feature representations.
Unfortunately, the steps involved in designing handcrafted features can be work-intensive and error-prone.
Motivated by the success of deep learning in other AI-relevant tasks based on images and speech \cite{lecun2015deep}, researchers have started to use deep models for representation learning in affect recognition.
As discussed in Section \ref{sec:deep-learning}, DNNs are more efficient at learning and representing the given functions, both statistically and regarding human involvement \cite{bengio2007scaling}.
Labeled data and computation time are limited and costly, incentivizing the adoption of such models in affect recognition.
Based on the previous discussion of challenges and model types, we identify three distinct ways in which deep learning is leveraged to learn useful feature representations for affect recognition tasks:

\begin{itemize}
	\item \textit{Learning spatial feature representations.}
	For images \cite{rifai2012disentangling}, short-term\footnote{For this study, we consider segments of up to 1 s as short-term.} image sequences \cite{gupta2017multi} and audio segments \cite{trigeorgis2016adieu}, DNNs and especially CNNs are used to learn spatial feature representations.
	\item \textit{Learning temporal feature representations.}
	To learn representations of the temporal dynamics found in audio \cite{mirsamadi2017automatic}, sequences of images \cite{kim2017multi}, and physiological measurements \cite{li2016emotion}, DNNs and especially RNNs are successfully applied.
	\item \textit{Learning joint feature representations} for multimodal data.
	In multimodal approaches, DNNs are leveraged to learn joint feature representations from multiple unimodal feature representations to accomplish feature-level fusion \cite{ringeval2015prediction}.
\end{itemize}

After learning such feature representations, the derived features (or \textit{deep features}) can be used as input for simple classification and regression methods, such as logistic regression and support vector regression (SVR).
In Section \ref{sec:state}, we give an in-depth breakdown of how deep learning models for human affect recognition are applied in practice.

\section{The state of the art}\label{sec:state}

\begin{table*}[!t]
\caption{Number of identified studies on human affect recognition by modality, and applications of deep learning.}
\label{tab:literature}
\begin{threeparttable}
	\centering
	\begin{tabularx}{\textwidth}{
	>{\hsize=.15\hsize}X
	>{\hsize=.2\hsize}X
	>{\hsize=.2\hsize}X
	>{\hsize=.2\hsize}X
	>{\hsize=.2\hsize}X
	>{\hsize=.1\hsize}X
	}
	\toprule
	\multicolumn{2}{c}{Modality} & \multicolumn{3}{c}{Learning deep feature representations} & Total\tnote{a} \\
	\cmidrule(lr){3-5}
	& & Spatial (Section \ref{sec:spatial}) & Temporal (Section \ref{sec:temporal})	& Joint (Section \ref{sec:joint}) &  \\
	\midrule
	\rule{0pt}{0ex}
	\multirow{2}{*}{Visual} & Facial expression & \nFERSpatial & \nFERTemporal & \nFERJoint & \nFER \\
							& Body movement		& \nBodySpatial & \nBodyTemporal & \nBodyJoint & \nBody \\
	\rule{0pt}{3ex}
	Auditory 				& Speech				& \nSERSpatial & \nSERTemporal & \nSERJoint & \nSER \\
	\rule{0pt}{3ex}
	\multirow{3}{*}{Physiological} & EEG			& \nEEGSpatial & \nEEGTemporal & \nEEGJoint & \nEEG \\
							& Peripheral			& \nPeriSpatial 	& \nPeriTemporal & \nPeriJoint & \nPeri \\
							& Other				& \nOtherSpatial & \nOtherTemporal & \nOtherJoint & \nOther \\
	\midrule
	Total\tnote{a}			&					& \nSpatial & \nTemporal & \nJoint & \nDLTotal \\
	\bottomrule
	\end{tabularx}
	\begin{tablenotes}
    	\item[a] Totals may not equal row/column sums due to overlaps between modalities and approaches.
    \end{tablenotes}
\end{threeparttable}
\end{table*}

Since around 2010, deep learning methods have started to fulfill crucial roles in human affect recognition systems.
Indeed, it is safe to state that they are currently driving most state-of-the-art results in this field.
Our goals are to (i) measure the adoption of deep learning in the field, and (ii) break down what specific functions are being fulfilled by DNNs in human affect recognition systems.

Because of these goals, we conducted a two-stage literature search\footnote{Databases searched: The ACM Digital Library, IEEE Xplore, SpringerLink, and Web of Science.} of studies on human affect recognition since 2010.
This search is restricted to studies that use sensor data of cues directly given by the human body\footnote{Since we only look at recognition of affective states, facial action unit detection and facial feature point detection are excluded. For SER, we restrict the scope to paralinguistic content, which is the main focus of research during the time frame covered by our review.}, i.e., facial expressions, movements, speech, and physiology.
Further affect recognition fields that focus on affect communicated through means other than the human body---such as generic images and videos \cite{chen2015learning}, music \cite{kim2010music}, \cite{schmidt2011learning}, and text \cite{calvo2013emotions}---are out of the scope of this review.
See \cite{cowie2001emotion}, \cite{pantic2003toward}, and \cite{zeng2009survey} for earlier reviews on human affect recognition.

\begin{figure}[!t]
\centering
\includegraphics[width=3in]{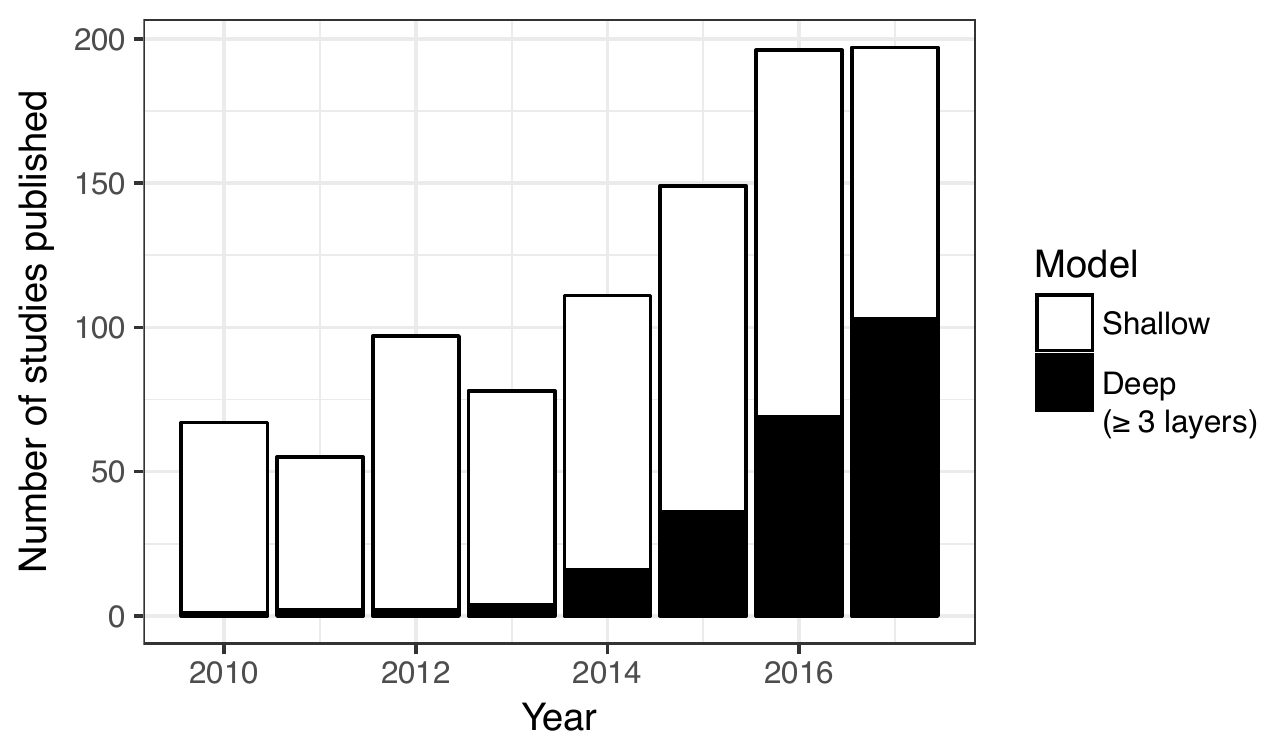}
\caption{Stage 1: Number of studies on affect recognition published by year, distinguished by their use of deep or shallow learning models.}
\label{fig:studies}
\end{figure}

Stage 1 of our search yielded a total of {\nMLTotal} studies.
Following the terminology introduced in Section \ref{sec:deep-learning}, these studies were then manually classified to indicate the use of shallow or deep models. 
Considering the temporal distribution of the studies, Fig. \ref{fig:studies} illustrates two developments:

\begin{itemize}
	\item Between 2010 and 2016\footnote{The search only contains partial data for 2017, hence 2017 is excluded here.}, there was an overall increase in the number of studies on human affect recognition 	({\avgRelIncr} year-on-year increase on average).
	\item Deep learning has gained considerable attention in this field since 2010: Up from one to two studies per year, it is being employed in {\prcDeepTotal} of studies in 2017---a {\avgRelIncrDL} average year-on-year increase in the number of published studies.
\end{itemize}

In Stage 2, we focus exclusively on the {\nDLTotal} studies found in our review that use deep learning for affect recognition.
They form the basis for the review in this section.
These studies were further classified by (i) the usage of deep learning according to the three ways introduced in Section \ref{sec:towards}, and (ii) the modality used as a basis for recognizing affect.
Table \ref{tab:literature} lays out the result of these classifications by listing the numbers of studies falling into each category.
Note that individual studies often use multiple modalities or apply deep learning in more than one way.

From Table \ref{tab:literature}, it is apparent that the application of DNNs for FER has attracted the most attention in the literature, featured in almost twice as many studies compared to SER, which is featured second most frequently.
Considering physiological signals, DNNs are most frequently applied in studies based on electroencephalography (EEG).

In the central part of Table \ref{tab:literature}, we can see that learning feature representations of spatial information is the most common application of DNNs in human affect recognition to date, especially in FER.
For SER in particular, learning of temporal feature representations with DNNs is an active research area.
A less active, but developing application area of DNNs in human affect recognition is learning joint feature representations for the purpose of early \textit{feature fusion} across different modalities.

\subsection{Learning spatial feature representations}\label{sec:spatial}

\begin{figure}[!t]
\centering
\includegraphics[width=3.2in]{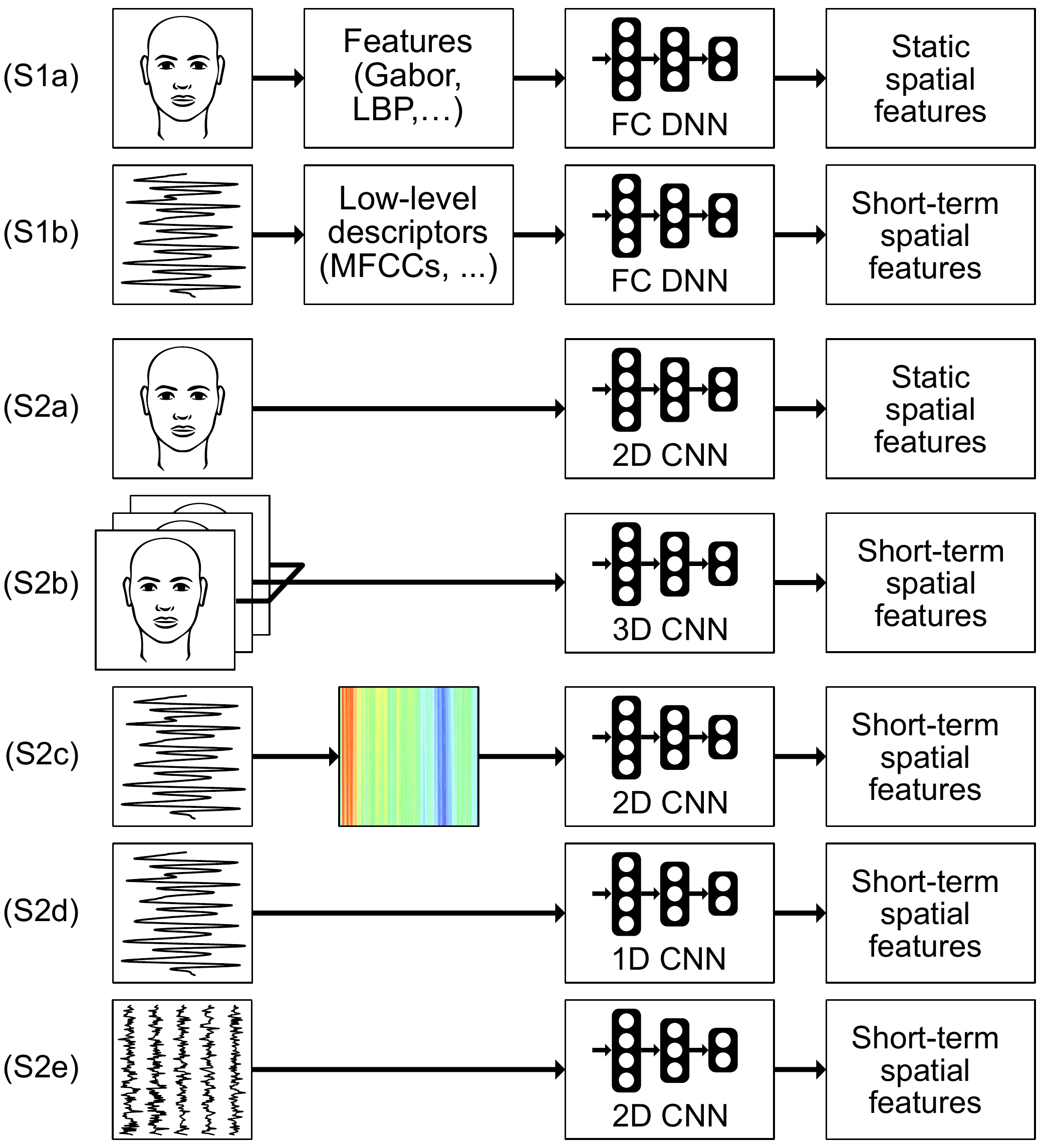}
\caption{Applications of deep learning for spatial feature learning with fully-connected DNNs (S1a--S1b) and CNNs (S2a--S2e).}
\label{fig:spatial}
\end{figure}

The goal in spatial feature learning is to learn expressive feature representations of data with spatial structure.
In practice, we find that deep architectures are frequently applied to exploit this characteristic in sensor recordings containing static and short-term cues of affective behavior---both visual imagery and short segments of audio and physiological data can be interpreted in this way.
Some approaches combine handcrafted features with deep architectures such as fully-connected DNNs (see Fig. \ref{fig:spatial}, S1a--S1b).
As discussed in Section \ref{sec:cnn}, the design of CNNs is based around the prior of spatial coherence.
Hence, CNNs are the most popular approach for learning spatial features (see Fig. \ref{fig:spatial}, S2a--S2e).  

\subsubsection{Learning spatial features for FER}\label{sec:spatial-FER}

Mehrabian \cite{mehrabian1968communication} famously posited that 55\% of the emotion conveyed in a message is perceived visually.
Indeed, FER is most prominently featured in our search results on spatial feature learning, as Table \ref{tab:literature} reveals.
Detailed surveys are available \cite{sariyanidi2015automatic}, \cite{corneanu2016survey}, giving an overview of the field in general.
In this section, we discuss the application of deep learning across {\nFERSpatial} studies to learn spatial features from images.
We distinguish between approaches using fully-connected DNNs (Fig. \ref{fig:spatial}, S1a), and CNNs (Fig. \ref{fig:spatial}, S2a--S2b).

Conventional approaches rely on handcrafted features to represent faces by their \textit{shape} or \textit{appearance}.
Shape representations use explicit knowledge about facial geometry to encode a given expression, such as the location of certain facial feature points.
The most common appearance features, such as local binary patterns (LBP), local phase quantization (LPQ), and histogram of oriented gradients (HoG), encode low-level texture information in local histograms.
Other methods of feature extraction include convolving the input with handcrafted Gabor filters and scale-invariant feature transform (SIFT).
As Sariyanidi et al. \cite{sariyanidi2015automatic} pointed out, such handcrafted features focus on low-level description of edge distributions.
While they provide robustness against illumination variations, they are less suitable for discrimination between high-level concepts such as facial features.
On the contrary, CNNs natively learn a hierarchy of features that builds from low-level to high-level representations.
While the first layer learns general concepts similar to Gabor filters \cite[ch.9.10]{goodfellow2016deep}, the last layers learn more specific concepts that tend to be semantically interpretable.
As a result, {\prcFERSpatialDeepBetter} of studies reporting direct comparisons find that deep spatial features outperform handcrafted spatial features for FER.

\textit{Learning spatial features from intermediate handcrafted features (S1a in Fig. \ref{fig:spatial}).}
As evident from Table \ref{tab:spatial}, especially during the initial adoption of DNNs in FER, handcrafted features and DNNs were combined in a sequential way.
In the first step, this approach extracts low-level handcrafted features from pixel values.
Appearance features such as LBP (e.g., \cite{levi2015emotion}, \cite{yao2016holonet}) and Gabor features (e.g., \cite{cheng2014deep}, \cite{lv2014facial}) are preferred for this approach.
Due to the reduced dimensionality, it is then feasible to apply fully-connected DBNs (e.g., \cite{cheng2014deep}) and SAEs (e.g., \cite{lv2014facial}, \cite{fadil2014multimodal}) for unsupervised learning of high-level features.

\begin{table}[!t]
\caption{Learning spatial feature representations: Number of studies adopting different approaches over time.}
\label{tab:spatial}
\centering
\begin{tabularx}{\columnwidth}{
>{\hsize=.1\hsize}X
>{\hsize=.1\hsize}X
>{\hsize=.1\hsize}X
>{\hsize=.1\hsize}X
>{\hsize=.1\hsize}X
>{\hsize=.1\hsize}X
>{\hsize=.1\hsize}X
>{\hsize=.1\hsize}X
}
\toprule
\multirow{2}{*}{Year} & \multicolumn{2}{c}{FC DNN} & \multicolumn{5}{c}{1D/2D/3D CNN} \\
\cmidrule(lr){2-3} \cmidrule(lr){4-8}
		& S1a	& S1b	& S2a	& S2b	& S2c	& S2d	& S2e	\\
\midrule
2012 	& -		& -		& 1		& -		& -		& -		& -		\\
2013 	& -		& 1		& 1		& -		& -		& -		& -		\\
2014 	& 3		& 1		& 4		& -		& 1		& -		& -		\\
2015 	& 3		& -		& 24		& 2		& 2		& -		& -		\\
2016 	& 2		& 2		& 42		& -		& 5		& 3		& 3		\\
2017 	& 3		& 2		& 57		& 9		& 11		& 9		& 1		\\
\midrule
Total 	& 11 	& 6		& 129	& 11	& 19		& 12		& 4		\\
\bottomrule
\end{tabularx}
\end{table}

\begin{figure}[!t]
\centering
\includegraphics[width=3.2in]{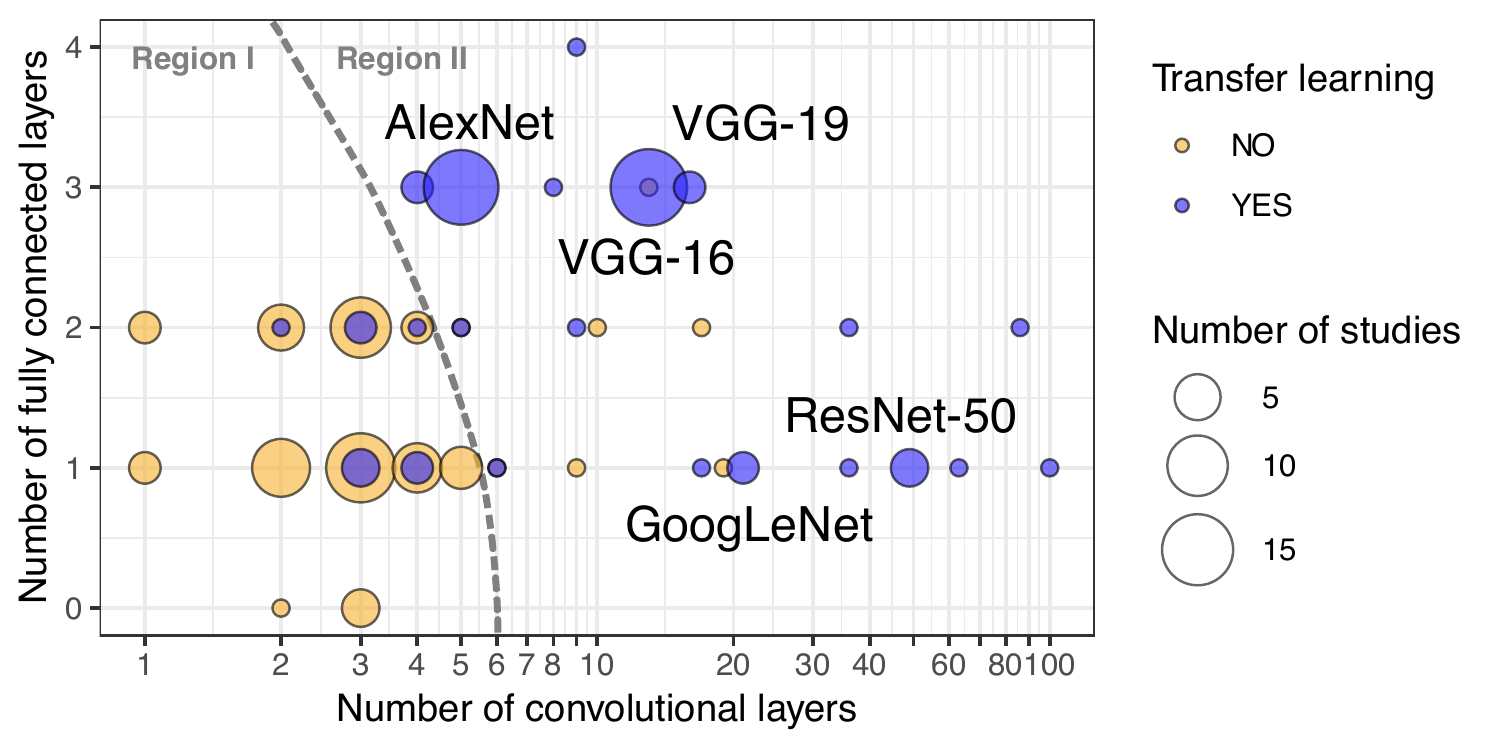}
\caption{Number of hidden layers in CNNs for spatial feature learning in FER.}
\label{fig:architecture-fer}
\end{figure}

\textit{Learning spatial features directly from 2D image with CNNs (S2a in Fig. \ref{fig:spatial}).}
CNNs are well suited to learn spatial features directly from image pixels.
As becomes clear from Tables \ref{tab:literature} and \ref{tab:spatial}, this approach dominates our search results.
Due to a lack of understanding why deep learning works well in practice \cite{zhang2017understanding}, and limited availability of labeled data, the challenge faced by researchers is choosing an appropriate architecture.
Fig. \ref{fig:architecture-fer} provides an overview of typical choices regarding the number of fully-connected and convolutional layers, and indicates whether transfer learning was used.
We can distinguish between two approaches:
Region I in the left of Fig. \ref{fig:architecture-fer} refers to architectures totalling six or less convolutional and fully-connected layers.
These CNN architectures, which are specifically designed for FER, make up {\prcFERSpatialRegionI} of the studies.
Most of these studies rely on smaller model sizes instead of transfer learning to avoid overfitting the relatively small number of examples.

Region II consists of architectures with more convolutional layers, and hence potential to learn more expressive features.
Many of these are existing architectures that have proven successful for other tasks such as object recognition.
Most frequently chosen are VGG Net \cite{simonyan2014very} ({\nVGG} studies) and AlexNet \cite{krizhevsky2012imagenet} ({\nAlexNet} studies).
AlexNet is the architecture that won the 2012 \ILSVRC \cite{russakovsky2015imagenet}.
It consists of five convolutional and three fully-connected layers, with a total of 60 million parameters.
VGG Net, an entry at the 2014 \ILSVRC, is a deeper CNN coming in variants of 16 and 19 layers.
Other choices include GoogLeNet \cite{szegedy2015going} (winner of {\ILSVRC} 2014), ResNet \cite{he2016deep} (winner of {\ILSVRC} 2015), and Tang's winning entry at ICML 2013's FER challenge \cite{tang2013deep}.
Most of these studies use additional datasets and transfer learning to avoid overfitting.

\textit{Learning short-term spatial features from image sequences with 3D CNNs (S2b in Fig. \ref{fig:spatial}).}
Building on the concept of spatial representation for a single 2D image, this approach interprets an image sequence as a spatio-temporal volume.
Standard CNNs can be extended to accept 3D volumes as input by increasing filter dimensionality to support spatio-temporal convolutions.
Such \textit{3D CNN} architectures \cite{ji20133d} are theoretically capable of learning spatio-temporal features such as motion of facial action units.
Since the number of model parameters and therefore the number of required examples increase with temporal depth of input sequences, the chosen number of frames is typically quite low\footnote{The reviewed studies used between 3 \cite{barros2015multimodal} and 16 \cite{zhang2017learning} frames.}, and 3D CNNs are limited to short-term sequences (up to 1 s).
The nature of this approach also requires that input sequences consist of a standardized number of frames.
This means that researchers need to downsample or interpolate videos \cite{gupta2017multi}.

\textit{Group-level FER.}
Group-level affect recognition is a subdiscipline of FER, where the goal is to assess the overall expression of all persons in an image.
It has been featured in the EmotiW competition in 2016 \cite{dhall2016emotiw} and 2017 \cite{dhall2017individual}.
For this purpose, spatial features are typically extracted for each person, and fused in some way---e.g., by considering multiple faces as a sequence and applying an LSTM \cite{li2016happiness}.

\textit{Complementarity of deep and handcrafted spatial features.}
In benchmarking experiments across various datasets, many studies found that features extracted with CNNs lead to higher recognition accuracies than handcrafted features (e.g., \cite{khorrami2015deep}, \cite{jung2015joint}, \cite{chen2016video}, \cite{khorrami2016deep}, \cite{breuer2017deep}, \cite{kim2017multi}, \cite{li2017multimodal}, \cite{gui2017curriculum}, \cite{lopes2017facial}).
However, several challenge-winning studies \cite{hu2017learning}, \cite{li2016happiness}, \cite{chen2017multimodal}, choose to use both handcrafted and deep features, e.g., by score-level \cite{hu2017learning} or model-level \cite{chen2017multimodal} fusion of separate handcrafted and deep models.
This suggests that deep and handcrafted features are complementary.
As of this writing, the findings of most studies doing related comparisons support this assumption \cite{ghazi2016automatic}, \cite{ding2016audio}, \cite{sun2016facial}, \cite{kaya2017video}, though in many cases a comparison is difficult as reported accuracies for well-performing fusion approaches include features from other modalities (see Section \ref{sec:joint}).
Only one study reported that deep and handcrafted features are not complementary \cite{xu2016video}.

\textit{Pre-processing to simplify the FER learning task.}
Instead of learning from unprocessed images, the majority of the reviewed studies apply some form of pre-processing to image data.
These steps reduce the amount of variation the model has to account for, and thus simplify the learning task.
\textit{Face cropping} is standard practice, and reported to increase model accuracy (e.g., from 54\% to 72\% in \cite{lopes2017facial}; see also \cite{li2016happiness}).
This involves the detection of the face and feature points, although some databases come with pre-detected or cropped faces (e.g., \cite{susskind2010toronto}, \cite{dhall2011static}, \cite{goodfellow2015challenges}).
\textit{Spatial normalization} techniques include face alignment and face frontalization:
Simple adjustment of face rotation and facial feature point alignment is reported to improve model accuracy (e.g., from 54\% to 62\% in \cite{lopes2017facial}; see also \cite{li2016happiness}, \cite{kaya2017video}).
More advanced face frontalization involving the approximation of 3D shape is useful when dealing with 3D head pose variation \cite{hassner2015effective}, \cite{rodriguez2017deep}.
\textit{Intensity normalization}, on the other hand, aims to normalize illumination related factors such as brightness and contrast, which can be an issue with images taken in the wild or across multiple databases.
Researchers report improvements in accuracy by applying intensity normalization (e.g., from 54\% to 57\% in \cite{lopes2017facial}; see also \cite{sun2016facial}).

\textit{Limited availability of labeled data.}
Since the number of labeled examples for FER remains relatively limited, large\footnote{When using CNNs for FER, researchers tend to resort to methods like transfer learning in Region II, see Fig. \ref{fig:architecture-fer}. Note however that the number of trainable parameters can be subject to many other factors.} models are likely to overfit the data \cite{kahou2015recurrent}, \cite{ng2015deep}, \cite{chen2016video}, \cite{tzirakis2017end}.
Several techniques are available to address this problem: (i) transfer learning, (ii) data augmentation, and (iii) architecture- and training choices promoting regularization, such as dropout.
In \textit{transfer learning}, the goal is to use additional corpora to learn generic visual descriptors that are found to be effective in improving initial model parameters \cite{sharif2014cnn}, and thus reduce overfitting.
Especially when adopting large models originally intended for object recognition, it is common to pre-train on a large-scale database such as \textit{ImageNet} \cite{deng2009imagenet} (14M annotated images).
Another database frequently chosen for pre-training is \textit{VGG-Face} \cite{parkhi2015deep} (2.6M face images).
Smaller, more relevant databases such as \textit{FER2013} \cite{goodfellow2015challenges} and \textit{CK+} \cite{lucey2010extended} are also frequently used for pre-training.
To make use of generic visual features, authors acquire large models pre-trained for object classification, "freeze" the lower-level layers, and fine-tune a selection of higher-level layers for affect recognition.
Improvements in accuracy are reported when following this approach (e.g., from 39\% to 42\% in \cite{chen2016video}; see also \cite{kaya2017video}, \cite{li2017multimodal}).

\textit{Data augmentation} is a technique whereby existing images or sequences are manipulated to reduce overfitting.
Researchers either use static rules to generate new examples for the same original label, or manipulate images randomly before training.
Such manipulations include horizontal flipping, cropping, rotation, translations, changes to color, brightness, and saturation, as well as scaling.
This way, researchers artificially increase the number of available examples or training epochs by a factor typically between 10 and 30 \cite{jung2015joint}, \cite{guo2016deep}, \cite{cai2016video}, and up to 300 \cite{ding2017facenet2expnet}.
Studies running experiments on this technique report accuracy improvements (e.g., from 79\% to 89\% in \cite{khorrami2015deep}; see also \cite{levi2015emotion}, \cite{lopes2017facial}).
\textit{Dropout} \cite{srivastava2014dropout} is a technique that reduces overfitting by randomly dropping out neurons during training, thus forcing the network to learn redundantly.
It is widely used in the reviewed studies---{\prcFERSpatialDropoutFC} report using dropout for fully-connected layers, and {\prcFERSpatialDropoutConv} report using it for convolutional layers.
Khorrami et al. \cite{khorrami2015deep} reported an increase in accuracy of 2.5\% after applying dropout to fully-connected layers.

\subsubsection{Learning spatial features for SER}\label{sec:spatial-SER}

Beyond spoken words, the acoustic properties of human speech are rich with information about the speaker, such as gender, age, and affect (see \cite{el2011survey}, \cite{anagnostopoulos2015features} for comprehensive reviews).
In this section, we focus on the {\nSERSpatial} studies that employ deep learning for spatial feature learning in SER.
The potential of replacing or complementing traditional short-term descriptors with DNNs in speech related classification tasks was pointed out as early as 2009 \cite{lee2009unsupervised}.
Especially CNNs are found useful in modeling speech features \cite{abdel2014convolutional}.
We distinguish between approaches using fully-connected DNNs (see Fig. \ref{fig:spatial}, S1b) and CNNs (see Fig. \ref{fig:spatial}, S2c--S2d).

Research has shown that short-term spectral, prosodic, and energy features of speech carry affective information \cite{eyben2016geneva}.
In conventional SER approaches, it is common practice to capture such properties using handcrafted features known as \textit{low-level descriptors} (LLDs).
LLDs are sampled from small overlapping audio segments or \textit{frames}; a common choice is a window size of 25 ms and a step size of 10 ms \cite{eyben2009openear}.
Most recent models use pre-defined sets of LLDs for spatial modeling.
Standard sets such as eGeMAPS \cite{eyben2016geneva} and ComParE \cite{schuller2013interspeech} typically include cepstral descriptors such as \textit{Mel-frequency cepstral coefficients} (MFCCs), energy-related descriptors such as \textit{shimmer} and \textit{loudness}, frequency-related descriptors such as \textit{pitch} and \textit{jitter}, and spectral parameters.
They can be extracted with software tools such as openSMILE \cite{eyben2010opensmile} and openEAR \cite{eyben2009openear}.
This review highlights how DNNs are used in the state of the art to complement and replace handcrafted LLDs.
Overall, {\prcSERSpatialDeepBetter} of studies reporting direct comparisons find that deep spatial features outperform handcrafted spatial features for SER.

\textit{Learning spatial features from handcrafted feature representations of speech (S1b in Fig. \ref{fig:spatial}).}
A limited number of early applications in SER were combinations of DNNs with handcrafted LLDs.
The idea was to use fully-connected DNNs to replace Gaussian Mixture Models (GMMs), which occupied the role of short-term modeling in the commonly used GMM-HMM architecture from ASR.
For example, Li et al. \cite{li2013hybrid} used a six-layer DNN to learn frame-level features from concatenated MFCCs of a sliding context window.
In their experiments, this yielded an accuracy improvement of more than 10\% over GMMs.

\textit{Learning spatial features from raw spectral representations of speech with CNNs (S2c in Fig. \ref{fig:spatial}).}
Many recent studies in SER leverage DNNs to avoid the step of handcrafted feature engineering (see Table \ref{tab:spatial}).
Mirsamadi et al. \cite{mirsamadi2017automatic} pointed out that most commonly used frame-level LLDs in SER can be derived from spectral representations of the raw speech signal.
Without any feature engineering, they were able to learn features similar to LLDs from the raw spectral representation of individual audio frames at 25 ms, leading to an accuracy increase of 4\%.
Such learned features can be shown to have similarities to handcrafted LLDs \cite{trigeorgis2016adieu}.

Spectrogram representations are computed using multiple frames and allow speech segments to be interpreted as 2D images.
They can be based on Fourier transform of the raw waveform (e.g., \cite{huang2014speech}, \cite{badshah2017speech}) or minimally hand-engineered on the log Mel-frequency cepstrum representation (e.g., \cite{zhang2017learning}, \cite{fayek2017evaluating}), which closer matches the characteristics of human auditory perception.
CNNs can be applied to directly learn features from such representations, which are typically between 250 ms and 1 s in length.
Since the suggested minimum time required to identify affect from speech is quoted in the literature as 250 ms \cite{wollmer2013lstm}, \cite{zhang2017learning}, the resulting features can directly be used for classification of short utterances (e.g., \cite{huang2014speech}, \cite{fayek2015towards}), or be regarded as short-term features (e.g., \cite{zhang2017learning}, \cite{zheng2015experimental}) for further temporal modeling as discussed in Section \ref{sec:temporal-SER}.

Most studies chose custom architectures of one to three convolutional layers and one to three fully-connected layers for this task.
Some considered using known architectures from object recognition \cite{ho2016emotion}, \cite{zhang2017learning}, \cite{badshah2017speech}.
Here, the authors regarded it as necessary to pre-train these larger models on the ImageNet database to avoid overfitting the relatively few labeled examples.

\textit{Learning spatial features from the raw waveform with CNNs (S2d in Fig. \ref{fig:spatial}).}
Feature learning directly from the raw waveform was proposed in 2011 \cite{jaitly2011learning}.
Since 2016 (see Table \ref{tab:spatial}), a number of studies have started applying this idea in SER.
CNNs can be applied to raw 1D audio (see Section \ref{sec:cnn}); indeed, all except one study in our review use CNNs for this task.
Trigeorgis et al. \cite{trigeorgis2016adieu} were the first to do so:
They used two convolutional layers for spatial modeling, almost doubling ground truth correlation over LLDs for arousal, and slightly improving for valence.
Bertero et al. \cite{bertero2017first} proposed one convolutional layer with 25 ms kernels, which achieved a 3\% accuracy improvement over LLDs.

\textit{Limited availability of labeled data.}
To avoid overfitting, both dropout \cite{srivastava2014dropout} and batch normalization \cite{ioffe2015batch} can help to achieve better regularization.
Dropout is applied in fully-connected (reported in {\prcSERSpatialDropoutFC} of studies, e.g., \cite{zhang2017multi}) and convolutional layers ({\prcSERSpatialDropoutConv} of studies, e.g., \cite{tzirakis2017end}).
Multiple studies have shown that transfer learning can improve model accuracy by leveraging additional sources of related knowledge (e.g., from other paralinguistic tasks \cite{gideon2017progressive}, various standard databases \cite{deng2013sparse}, and different affect representations \cite{zhang2017multi}).
SoundNet \cite{aytar2016soundnet}, a 1D CNN trained with unlabeled video, has been shown to perform well in SER even without fine-tuning \cite{pini2017modeling}, and was featured in a challenge-winning submission \cite{chen2017multimodal}.
Semi-supervised learning can give access to knowledge contained in unlabeled datasets \cite{deng2017universum}.
Knowledge transfer from domains like music \cite{coutinho2014transfer} and visual object recognition \cite{ho2016emotion}, \cite{zhang2017learning} is also possible.
Data augmentation to artificially increase the dataset size is used less often than for FER;
notable examples include the addition of Gaussian noise \cite{wollmer2013lstm}, different sampling frequencies \cite{fayek2015towards}, and modified playback speed \cite{aldeneh2017using}.

\subsubsection{Learning spatial features from physiology}\label{sec:spatial-physio}

As early as 2001, physiological responses were shown to convey information about affective states in machine learning \cite{picard2001toward}.
Candidates include measures of the peripheral physiology via electrocardiography (ECG), electrodermal activity (EDA), and brain activity via EEG.
However, affective computing research initially focused mostly on FER and SER, partially due to a lack of interest and inconvenient sensors \cite{pantic2003toward}.
More recently, interest in physiological affect recognition is seeing a resurgence \cite{fairclough2008fundamentals}, owed in part to the capabilities of modern, portable monitoring devices \cite{wilhelm2010emotions}.

As part of our review, we identified only four studies using 2D CNNs to learn spatial features from EEG data (see S2e in Fig. \ref{fig:spatial}).
All achieved accuracy improvements over handcrafted approaches, but a lack of training data was also mentioned \cite{yanagimoto2016recognition}, \cite{zhang2017spatial}.
For example, Yanagimoto and Sugimoto \cite{yanagimoto2016recognition} divided the raw 16-channel EEG data into 1s segments and used a seven-layer CNN with 10 ms kernels on the first layer, leading to accuracy improvements of over 20\%.
Similar to some work in SER, Li et al. \cite{li2016emotion} considered a spectrogram representation of the EEG signal at a frame size of 1s.
Another way to learn spatial features from the EEG signal is to reflect it as a 2D map representing the location of the electrodes on the head \cite{li2016implementation}.

\subsubsection{Takeaways for spatial feature learning}

\begin{figure}[!t]
\centering
\includegraphics[width=3in]{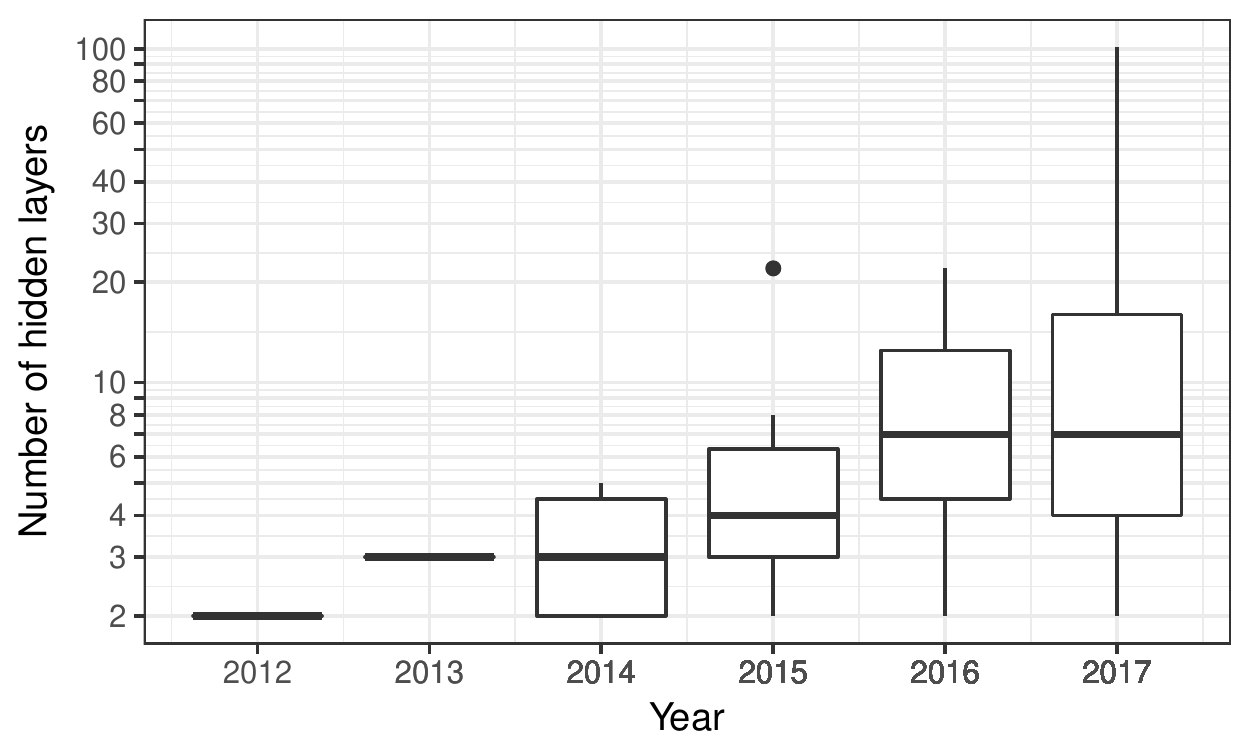}
\caption{Number of hidden layers used for spatial feature learning in FER.}
\label{fig:layers}
\end{figure}

\begin{itemize}
	\item Deep spatial features lead to higher accuracies than handcrafted spatial features. Out of {\nSpatialComparison} studies that reported comparisons, {\prcSpatialDeepBetter} support this finding.
	\item However, in contrast to fields such as object- and speech recognition, both are often found to be complementary in affect recognition as of this writing. This suggests that the full potential of deep learning in affect recognition may not have been seen yet.
	\item CNNs are the most widely used architecture for spatial feature learning (91\% of the studies in our review; see Table \ref{tab:spatial}). Instead of using spectrogram representations, recent research starts to apply CNNs directly to raw speech and physiological data.
	\item To achieve higher accuracy, research strives towards ``deeper'' models (Fig. \ref{fig:layers} illustrates this for FER), but runs into the problem of overfitting. This is the main challenge for current research.
\end{itemize}

\subsection{Learning temporal feature representations}\label{sec:temporal}

\begin{figure}[!t]
\centering
\includegraphics[width=3.2in]{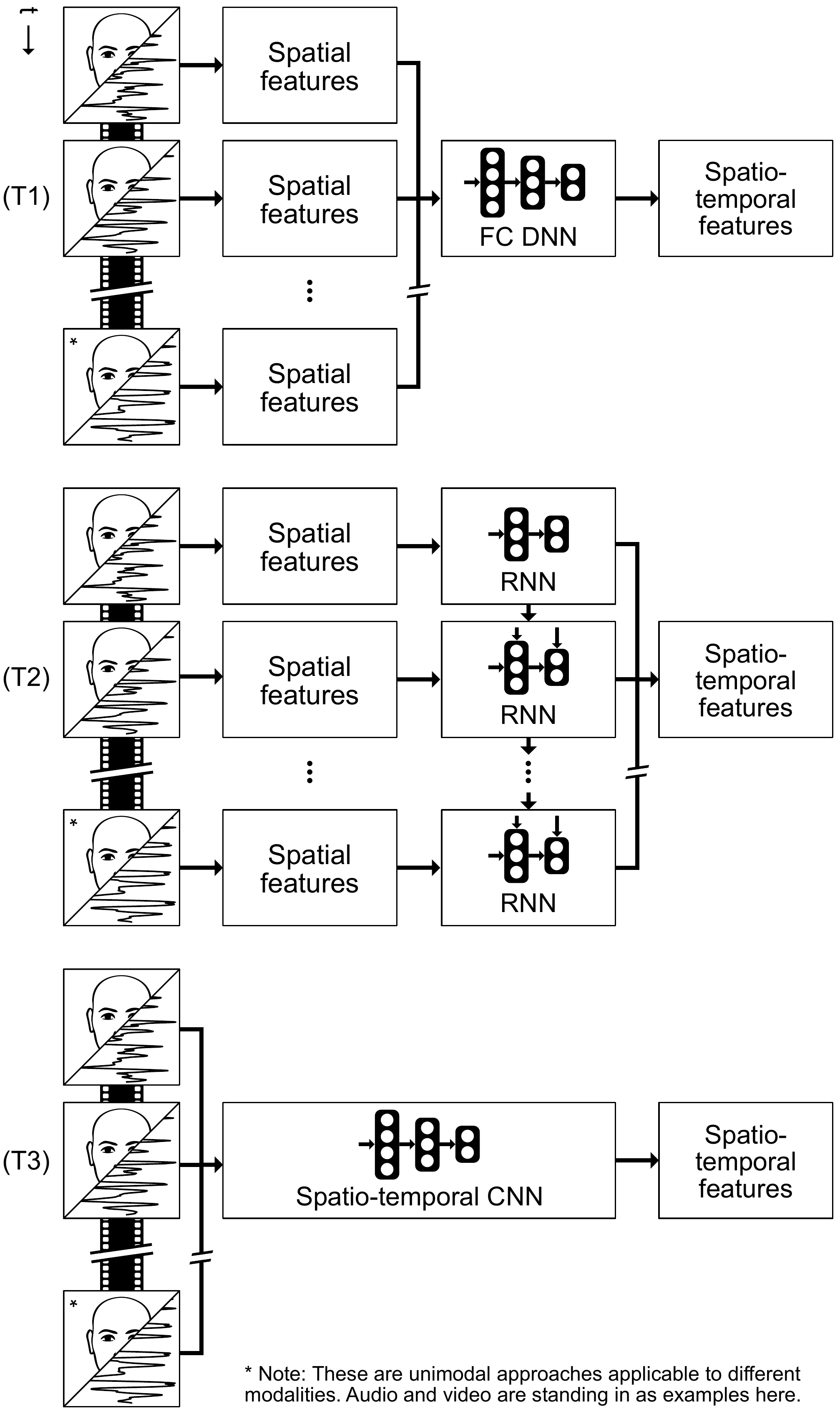}
\caption{Applications of deep learning for temporal feature learning with fully-connected DNNs (T1), RNNs (T2), and CNNs (T3).}
\label{fig:temporal}
\end{figure}

When learning from sequences, the goal is to learn feature representations that capture temporal dynamics \cite{lipton2015critical}.
This allows models to consider the temporal variation of spatial characteristics in sensor data (e.g., in SER \cite{anagnostopoulos2015features} and video-based FER \cite{kahou2015recurrent}).
As discussed in Section \ref{sec:deep-learning}, both CNNs and RNNs provide architectures that can learn representations of sequences of data.
We found that the existing architectures---spanning all studies and modalities---can be classified into one of three approaches illustrated in Fig. \ref{fig:temporal}: (T1) fully-connected DNNs for learning spatio-temporal features from aggregated frame-level spatial features, (T2) RNNs for global temporal modeling based on frame-level spatial features, and (T3) CNNs for local temporal modeling.

\begin{table}[!t]
\caption{Learning temporal feature representations: Number of studies adopting different approaches over time.}
\label{tab:temporal}
\begin{threeparttable}
\centering
\begin{tabularx}{\columnwidth}{
>{\hsize=.2\hsize}X
>{\hsize=.1\hsize}X
>{\hsize=.1\hsize}X
>{\hsize=.1\hsize}X
>{\hsize=.1\hsize}X
>{\hsize=.1\hsize}X
>{\hsize=.1\hsize}X
>{\hsize=.1\hsize}X
>{\hsize=.1\hsize}X
>{\hsize=.1\hsize}X
}
\toprule
\multirow{2}{*}{Year} & \multicolumn{3}{c}{FC DNN (T1)\tnote{a}} & \multicolumn{3}{c}{RNN (T2)\tnote{a}} & \multicolumn{3}{c}{CNN (T3)\tnote{a}} \\
\cmidrule(lr){2-4} \cmidrule(lr){5-7} \cmidrule(lr){8-10}
	 	& V & A  & P  & V  & A  & P & V & A  & P \\
\midrule
2010 	& - & -	 & -  & -  & 1  & - & - & -  & - \\
2011 	& -	& 1	 & -	  & 1  & 1  & - & - & -  & - \\
2012 	& -	& -	 & -	  & 1  & 1  & - & - & -  & - \\
2013 	& -	& -	 & -	  & -  & -  & - & - & -  & 1 \\
2014 	& - & 2  & 2  & 2  & 3  & 1 & - & -  & - \\
2015 	& 2	& 3  & 2  & 1  & -  & - & 1 & 1  & - \\
2016 	& 3	& 10 & 4  & 10 & 5  & 2 & - & 5  & 2 \\
2017 	& 1	& 11 & 6  & 15 & 13 & 1 & 3 & 12 & 1 \\
\midrule
Total 	& 6 & 27 & 14 & 30 & 24 & 4 & 4 & 18 & 4 \\
\bottomrule
\end{tabularx}
\begin{tablenotes}
\item[a] Modalities: V = Visual (FER, Body), A = Audio (SER), P = Physio (EEG, Peripheral, Other).
\end{tablenotes}\end{threeparttable}
\end{table}

\subsubsection{Learning temporal features for FER}\label{sec:temporal-FER}

A straightforward approach to derive sequence-level features from video data is to first extract high-level spatial features (such as facial characteristics, see Fig. \ref{fig:au}) from individual frames, and then aggregate these in some way. 
This can be achieved by simple feature pooling strategies such as mean pooling, max pooling, or feature concatenation.
However, such strategies typically ignore most of the temporal variation in the sequence, which may contain valuable contextual information.
Well-designed models seek to further exploit such information.
To some extent, this is possible with common handcrafted features:
Appearance features can be extended for spatio-temporal representation by considering a third orthogonal plane \cite{ghazi2016automatic}, \cite{kaya2017video}.
With DNNs, temporal modeling capabilities in FER can further be improved:
{\prcFERTemporalDeepBetter} of studies reporting comparisons with handcrafted approaches find that deep temporal features perform better for FER.

\textit{Learning spatio-temporal features from aggregated frame-level spatial features (T1 in Fig. \ref{fig:temporal}).}
In some cases, fully-connected DNNs are applied to achieve dimensionality reduction on high-dimensional spaces of aggregated handcrafted features.
For example, Zhang et al. \cite{zhang2016automatic} and Ranganathan et al. \cite{ranganathan2016multimodal} used fully-connected DBN models to learn from aggregated facial feature point trajectories, both improving recognition accuracies over shallow aggregation strategies.			

\textit{Global temporal modeling with RNN based on frame-level spatial features (T2 in Fig. \ref{fig:temporal}).}
The properties of RNNs, as discussed in Section \ref{sec:deep-learning}, make them well-suited to model the temporal variation of frame-level spatial features.
This approach first extracts high-level spatial features from each face image, which are then considered as sequential input to the RNN.
Advantages of this approach include the ability to process long sequences, and the possibility of both sequence-level and continuous frame-level affect recognition on image sequences of arbitrary length.
Early on, RNNs were used to learn temporal context from handcrafted spatial features such as coordinates of facial feature points \cite{nicolaou2011continuous}, optical flow \cite{wollmer2013lstm}, and LBP \cite{wei2014multimodal}.

More recent studies combine RNNs with deep methods for spatial feature learning discussed in Section \ref{sec:spatial-FER}, by adopting deep features from the last layer of a CNN trained for affect recognition (e.g., \cite{kahou2015recurrent}, \cite{brady2016multi}, \cite{kim2017multi}).
We see both CNN-RNN (e.g., \cite{brady2016multi}, \cite{khorrami2016deep}) and CNN-LSTM (e.g., \cite{tzirakis2017end}, \cite{rodriguez2017deep}, \cite{kim2017multi}) architectures, with CNN-LSTM being the more frequent choice among the reviewed studies.
Global temporal modeling is found to lead to improved accuracies when compared with simpler methods such as pooling of spatial features (e.g., \cite{yan2016multi}, \cite{khorrami2016deep}, \cite{tzirakis2017end}).
A disadvantage of most CNN-LSTM implementations is that training occurs in a disconnected way:
The CNN is trained on frame-level, specifically for static spatial affect recognition.
Hence, the extracted features are not necessarily optimal for further temporal context learning by the RNN.
End-to-end training of the entire CNN-LSTM system addresses this problem, and can lead to accuracy improvements \cite{tzirakis2017end}.

\textit{Local temporal modeling with 3D CNN (T3 in Fig. \ref{fig:temporal}).}
When using 3D CNN for spatio-temporal modeling of image sequences as discussed in Section \ref{sec:spatial-FER}, the line between spatial and temporal representation learning can be blurred.
While this approach is typically limited to very short sequences, with further pooling steps necessary to derive sequence-level labels (e.g., \cite{barros2015multimodal}, \cite{zhang2017learning}), in some cases spatio-temporal features can be derived for entire (short) sequences.
For example, Gupta et al. \cite{gupta2017multi} used a variant called \textit{slow fusion} \cite{karpathy2014large}, which treats the time domain like a spatial domain, progressively learning low-level to high-level temporal features.
As the amount of parameters required due to the temporal depth of the input is effectively reduced, this allows for more input frames.

\subsubsection{Learning temporal features from body movement}

Besides facial expression, body movement and gestures are other means of expressing affect visually \cite{picard1997affective}.
In the reviewed studies, spatial features representing such movements are extracted using skeletal and shoulder tracking.
For example, Ranganathan et al. \cite{ranganathan2016multimodal} and Kaza et al. \cite{kaza2016body} used the approach illustrated in Fig. \ref{fig:temporal} (T1), to learn spatio-temporal features from statistics of skeletal tracking point trajectories.
Shoulder cues were used by Nicolaou et al. \cite{nicolaou2011continuous} in the RNN approach illustrated in Fig. \ref{fig:temporal} (T2).
In comparison with facial expression and speech, they were found to be less expressive for prediction of both arousal and valence.

\subsubsection{Learning temporal features for SER}\label{sec:temporal-SER}

To derive fixed-length features at the utterance level, SER models traditionally aggregate LLDs by \textit{high-level statistical functionals} (HSFs) such as mean and standard deviation.
Standard sets of HSFs and LLDs are given in eGeMAPS \cite{eyben2016geneva} and ComParE \cite{schuller2013interspeech}.
HMMs have long served as a standard choice for further modeling of temporal variation in speech signals, especially in ASR \cite{gales2008application}, but also in SER \cite{li2013hybrid}.
More recently, RNNs have emerged as a preferred way of modeling the sequential aspect of speech \cite{graves2013speech}.
In particular, {\prcSERTemporalDeepBetter} of studies reporting comparisons with handcrafted approaches find that deep temporal features perform better for SER.
This review highlights how DNNs can complement or replace both HSFs and HMMs for learning temporal representations in SER.

\textit{Learning from aggregated frame-level features (T1 in Fig. \ref{fig:temporal}).}
Since there is no consensus in the literature over a "universal" handcrafted feature set with superior performance \cite{eyben2016geneva}, many recent studies have applied a "brute-force" approach, resulting in a large number of features per utterance.
This number varies from several hundred \cite{schuller2009interspeech} to several thousand \cite{schuller2012interspeech}, \cite{schuller2013interspeech}, depending on the employed LLDs and HSFs.
DNNs can be integrated to learn more high-level representations of these handcrafted spatio-temporal feature spaces.
Studies aiming to reduce feature space dimensionality with deep learning almost exclusively use fully-connected DNNs, consisting of two to four hidden layers.
It is common to initialize model parameters layer-wise via unsupervised pre-training as RBMs (e.g., \cite{kim2013deep}, \cite{kahou2016emonets}), or AEs (e.g., \cite{fadil2014multimodal});
subsequently, a Softmax classification layer is added for supervised fine-tuning.
Alternatively, DBNs or SAEs can serve as feature extractors for classification via support vector machine (e.g., \cite{deng2013sparse}).
Dimensionality reduction of handcrafted features with DNNs can lead to improvements in accuracy over various databases \cite{stuhlsatz2011deep}.

\textit{Global temporal modeling based on frame-level features with RNNs (T2 in Fig. \ref{fig:temporal}).}
In this approach, an utterance-level RNN models the temporal variation of frame-level features.
Most straightforwardly, LLDs can directly be fed into the RNN at the frame level (e.g., \cite{coutinho2014transfer}, \cite{lingenfelser2016asynchronous}, \cite{mirsamadi2017automatic}).
Depending on the nature of the source audio, it can be beneficial to apply HSFs to frame-level LLDs according to a sliding window before applying the RNN \cite{wei2014multimodal}, \cite{ringeval2015prediction}.
One study suggested that a smaller window size (2s) could be the best choice \cite{ringeval2015prediction}.
In general, the addition of RNN for temporal modeling is associated with an increase in model accuracy (e.g., \cite{wollmer2013lstm}, \cite{ding2016audio}, \cite{lingenfelser2016asynchronous}).
When dealing with dimensional labels, this allows learning features at the frame level.
Here, LSTM is found to outperform state-of-the-art techniques like SVR (e.g., \cite{nicolaou2011continuous}, \cite{wei2014multimodal}).

Since 2016, {\nSERCNNLSTM} studies have explored combining deep spatial features and RNN-based temporal feature learning.
For spectrogram-based spatial features, Lim et al. \cite{lim2016speech} found that the CNN-LSTM architecture yields the best result on Emo-DB \cite{burkhardt2005database}.
Applications of similar architectures based on the raw waveform have also been attempted \cite{trigeorgis2016adieu}, \cite{tzirakis2017end}, showing that end-to-end learning can outperform shallow models.
Overfitting is still a problem for this approach due to the large number of model parameters and limited dataset sizes.
Mirsamadi et al. \cite{mirsamadi2017automatic} found that model performance is slightly lower with joint learning of both short-term spatial features and temporal context on the IEMOCAP dataset \cite{busso2008iemocap}, while both improve model performance when applied independently.

\textit{Local temporal modeling with CNNs (T3 in Fig. \ref{fig:temporal}).}
When CNNs are used for modeling speech, they typically combine spatial modeling in the short term with temporal modeling of longer segments or entire utterances.
The kernels of higher-level (i.e., second or third) convolutional layers can often be interpreted as learning temporal structure based on spatial features learned by the kernels in the first layer (see Section \ref{sec:spatial-SER}).
For example, Trigeorgis et al. \cite{trigeorgis2016adieu} performed pooling across time after learning spatial characteristics from the raw signal in the first layer, and added a second layer with 500 ms kernels to learn temporal characteristics.
Similarly, Zhang et al. \cite{zhang2017learning} used an AlexNet to add increasingly more temporal context to learned feature representations. 
It is worth noting that while some studies directly used CNN features for affect prediction \cite{aldeneh2017using}, \cite{fayek2017evaluating}, others combined local temporal modeling with global temporal modeling via RNN \cite{tzirakis2017end}, \cite{lim2016speech}, or pooling approaches \cite{zhang2017learning}.

\subsubsection{Learning temporal features from physiological data}\label{sec:physio}

\textit{Learning from handcrafted features with fully-connected DNNs (T1 in Fig. \ref{fig:temporal}).}
As highlighted in Table \ref{tab:temporal}, approach T1 is the primary application of DNNs for feature learning from physiological data.
For this purpose, fully-connected DNNs are initialized by iterative training and stacking of unsupervised models such as RBMs or AEs \cite{jirayucharoensak2014eeg}, and applied to functionals of frame-level spatial features.
Typical for EEG are handcrafted features derived from the frequency domain, such as power spectral density (PSD) coefficients of different frequency bands.
Zheng et al. \cite{zheng2014eeg} found that DBNs can improve recognition accuracy of models based on differential entropy features.
Similarly, Xu and Plataniotis \cite{xu2016affective} showed that DBNs can build on PSD features to outperform state-of-the-art methods on the DEAP dataset.
Deep learning has also been used as part of ensemble methods \cite{mehmood2017optimal}, and in the form of Echo State Networks \cite{bozhkov2016learning} for dimensionality reduction of handcrafted EEG features.
Yin et al. \cite{yin2017recognition} used stacked autoencoders (SAEs) to learn high-level representations from various peripheral sensors including skin temperature and blood volume pressure, improving the state-of-the-art by 5\%.

\textit{Learning temporal context from spatial features with RNNs (T2 in Fig. \ref{fig:temporal}).}
RNNs have been used to learn temporal context from EEG features to improve recognition accuracies \cite{zhang2017spatial}, \cite{li2016emotion}.
Brady et al. \cite{brady2016multi} found that learning temporal context with LSTM and handcrafted features leads to improvements over shallow baseline models.
Ringeval et al. \cite{ringeval2015prediction} used a similar approach.
They found that while the given physiological signal has lower predictive power than audiovisual signals, both are complementary.

\textit{Learning spatio-temporal representations from raw data with CNNs (T3 in Fig. \ref{fig:temporal}).}
A limited number of {\nPhysioTiii} studies have attempted to learn spatio-temporal features directly from raw physiological data for discrimination between affective states.
Yanagimoto and Sugimoto \cite{yanagimoto2016recognition} used a CNN on raw 16-channel EEG data to differentiate between positive and negative affective states, which is shown to outperform shallow models based on common features.
Similar results are reported when learning from intermediate representations based on differential entropy \cite{li2016emotion}.
Martinez et al. \cite{martinez2013learning} were the first to learn deep features directly from the peripheral physiology.
For this purpose, they used CNNs trained in an unsupervised way via AEs to learn features from raw blood volume pulse and skin conductance signals, which outperformed models based on handcrafted features.

\subsubsection{Takeaways for temporal feature learning}

\begin{itemize}
	\item Deep temporal features lead to higher accuracies than handcrafted temporal features. Out of {\nTemporalComparison} studies that reported comparisons, {\prcTemporalDeepBetter} support this finding.
	\item While CNNs are well suited for local temporal modeling, RNNs are found to be useful for global temporal modeling of affect.
	\item Since 2015, there are studies using deep learning for both spatial and temporal feature learning.
	\item We are starting to see studies implementing end-to-end training for such models \cite{trigeorgis2016adieu}, \cite{tzirakis2017end}, however in this setting the problem of limited labeled data becomes especially noticeable \cite{mirsamadi2017automatic}, \cite{pini2017modeling}.
\end{itemize}

\subsection{Learning joint feature representations}\label{sec:joint}

\begin{table}[!t]
\caption{Learning joint feature representations: Number of studies adopting different approaches over time.}
\label{tab:joint}
\begin{threeparttable}
\centering
\begin{tabularx}{\columnwidth}{
>{\hsize=.1\hsize}X
>{\hsize=.1\hsize}X
>{\hsize=.1\hsize}X
>{\hsize=.1\hsize}X
>{\hsize=.1\hsize}X
>{\hsize=.1\hsize}X
>{\hsize=.1\hsize}X
>{\hsize=.1\hsize}X
>{\hsize=.1\hsize}X
}
\toprule
\multirow{2}{*}{Year} & \multicolumn{4}{c}{FC DNN (J1)\tnote{a}} & \multicolumn{4}{c}{RNN (J2)\tnote{a}} \\
\cmidrule(lr){2-5} \cmidrule(lr){6-9}
	 	& VA	& VAP	& VP		& PP	& VA		& VAP	& VP		& PP	\\
\midrule
2011 	& -		& -	 	& -	  	& - 	& 1		& -		& -		& -		\\
2012 	& -		& -		& -	  	& -		& 1		& -		& -		& -		\\
2013 	& 1		& -		& -	  	& -		& -		& -		& -		& -		\\
2014 	& - 	& -		& -  	& -		& -		& 1		& -		& -		\\
2015 	& 2		& 1		& 1  	& -		& -		& -		& -		& -		\\
2016 	& 2		& 1		& -  	& 1		& 1		& -		& -		& -		\\
2017 	& 3		& -		& -  	& 1		& 4		& -		& -		& -		\\
\midrule
Total 	& 8 	& 2		& 1		& 2		& 7		& 1		& 0 		& 0		\\
\bottomrule
\end{tabularx}
\begin{tablenotes}
\item[a] Modalities: V = Visual (FER, Body), A = Audio (SER), P = Physio (EEG, Peripheral, Other).
\end{tablenotes}\end{threeparttable}
\end{table}

It is generally accepted in the literature that multimodal (e.g., audiovisual) sensor combinations have complementary effects and thus may increase model accuracy \cite{zeng2009survey}.
The challenge in joint multimodal feature learning is how and at what stage to fuse data from multiple modalities.
This challenge is complicated by the high dimensionality of raw data, differing temporal resolutions, and differing temporal dynamics across modalities.
Surveys on the general problem of sensor fusion \cite{lingenfelser2011systematic} and specifically on fusion for affect recognition \cite{dmello2015review}, \cite{poria2017review} are available.

Fusion can be achieved at early model stages close to the raw sensor data, or at a later stage by combining independent models.
In early or \textit{feature-level} fusion, features are extracted independently and then concatenated for further learning of a joint feature representation; this allows the model to capture correlations between the modalities.
Late or \textit{decision-level} fusion aggregates the results of independent recognition models.
To date, the literature generally reports that decision-level fusion works better for affect recognition given the datasets and models currently used \cite{ringeval2015prediction}.
While decision-level fusion typically only involves simple score weighing, feature-level fusion is a representation learning task that may benefit from deep learning.
Here, we report on the approaches of {\nJoint} studies that use deep learning for joint feature learning from multimodal data.

\begin{figure}[!t]
\centering
\includegraphics[width=3.2in]{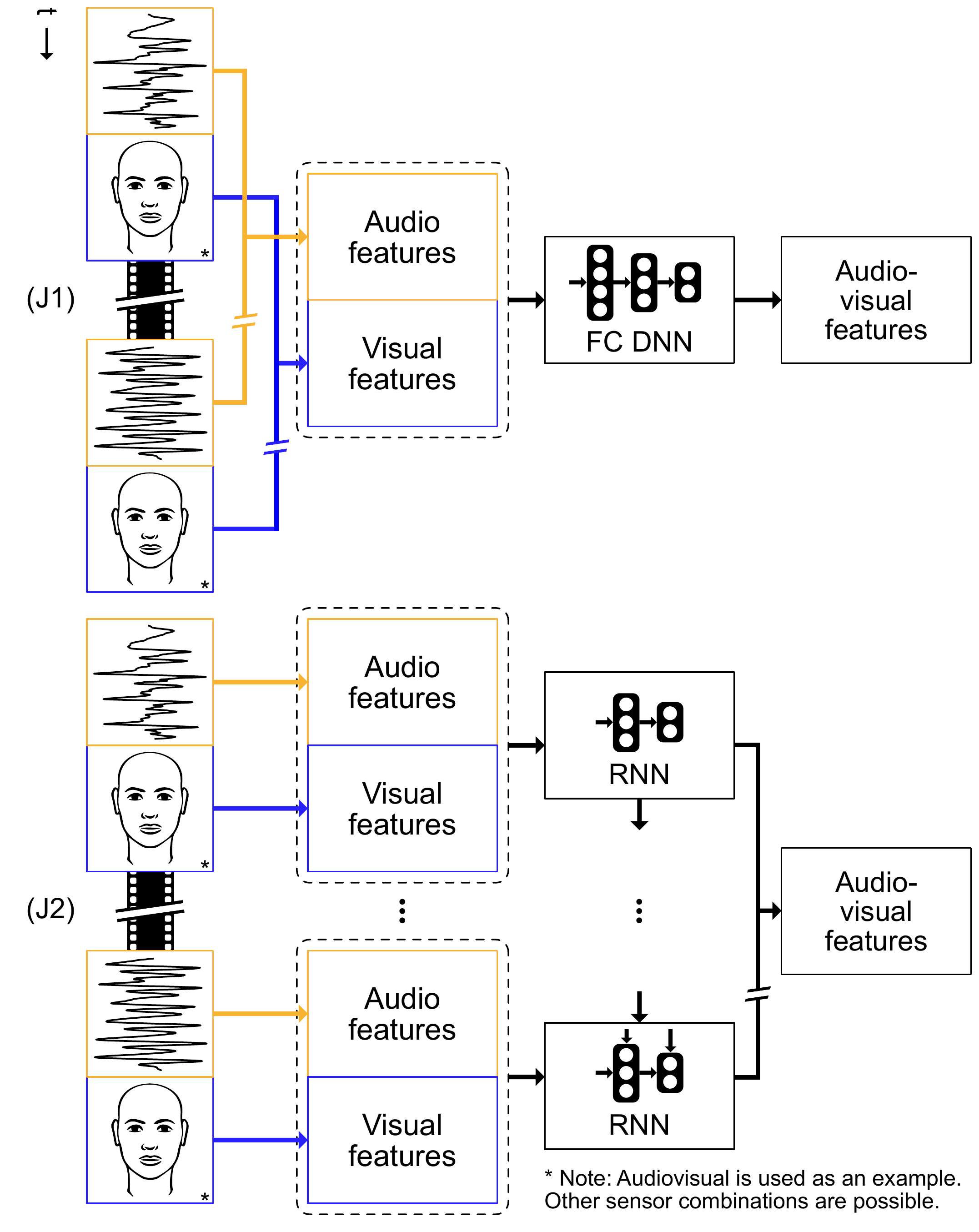}
\caption{Applications of deep learning for joint multimodal feature learning with fully-connected fusion DNNs (J1) and fusion RNNs (J2).}
\label{fig:joint}
\end{figure}

\subsubsection{Learning joint features with audiovisual data}

The most common sensor combination found in {\nAudioVisual} studies involves facial expressions and speech.

\textit{Feature-level fusion with fully-connected DNNs (J1 in Fig. \ref{fig:joint}).}
In this approach, joint feature representations are learned without considering the temporal context for fusion.
For both modalities, video-level features are extracted using FER and SER methods that may involve both handcrafted and deep features (see Sections \ref{sec:spatial} and \ref{sec:temporal}).
A fully-connected DNN, typically initialized via unsupervised pre-training, then learns a high-level joint feature representation of both modalities as an improvement over ``shallow'' feature fusion.
Kim et al. \cite{kim2013deep} and others (e.g., \cite{ranganathan2016multimodal}, \cite{zhang2016automatic}) demonstrated how this can be achieved with DBNs.
This approach is feasible especially in cases where the goal is to label each video with one affective state.
Alternatively, joint feature representations can be learned at the frame level, and then aggregated to the video level:
Zhang et al. \cite{zhang2017learning} used a DBN to fuse frame-level audiovisual features learned independently via CNNs; the learned features are average-pooled for classification at the video level and lead to an improvement over state-of-the-art methods.

\textit{Feature-level fusion with RNNs (J2 in Fig. \ref{fig:joint}).}
Especially when predictions are required at the frame level for dimensional affective states, feature-level fusion could benefit by taking into account the temporal context.
Modeling via RNNs makes this possible, potentially improving model robustness and helping to deal with temporal lags between modalities \cite{lingenfelser2016asynchronous}.
Initial studies reported that dynamic feature fusion can lead to performance improvements compared to simpler fusion strategies \cite{lingenfelser2016asynchronous}.
However, several other studies based on handcrafted features found that decision-level fusion on top of individual LSTM models leads to better performance \cite{ringeval2015prediction}, \cite{wollmer2013lstm}.
Learning from raw audiovisual data with two CNNs, Tzirakis et al. \cite{tzirakis2017end} used a two-layer LSTM network for feature fusion, which was found to outperform the state of the art.

\subsubsection{Learning joint features with physiological data}

A small number of {\nAVP} studies combined the audiovisual and physiological (AVP) modalities.
Ranganathan et al. \cite{ranganathan2016multimodal} demonstrated the feasibility of learning joint feature representations of AVP sensor data with approach J1 and a DBN, but do not compare the performances of different modality combinations.
Ringeval et al. \cite{ringeval2015prediction} used approach J2 with the AVP modalities and LSTM.
They concluded that in feature-level fusion, ECG data helps for prediction of valence, but not arousal.

Feature-level fusion can also be based solely on physiological measurements.
Yin et al. \cite{yin2017recognition} successfully used a fusion SAE to aggregate handcrafted features from several different sensors.
Similarly, Liu et al. \cite{liu2016emotion} used handcrafted features derived from EEG and eye tracking as input into a SAE. 
Both studies found that the representation learned through feature-level fusion leads to improved accuracy over individual modalities.

\subsubsection{Takeaways for joint feature learning}

\begin{itemize}
	\item Joint feature learning is most commonly applied to audiovisual fusion (see Table \ref{tab:joint}).
	\item To date, there is no consensus whether feature-level fusion with deep learning leads to superior accuracy over simple decision fusion. Out of {\nJointComparison} studies that reported comparisons, only {\prcJointDeepBetter} find that it does.
	\item While dimensional models of affect are only used in {\prcDimSpatial} of spatial and {\prcDimTemporal} of temporal feature learning studies, they are employed in {\prcDimJoint} of studies on joint feature learning.
\end{itemize}

\subsection{Databases and competitions}\label{sec:databases}

\def\sequence{
\begin{tikzpicture}[scale=0.25]
\filldraw[fill=white, draw=black] (0.2,0.2) rectangle (0.9,0.9);
\filldraw[fill=white, draw=black] (0.1,0.1) rectangle (0.8,0.8);
\filldraw[fill=white, draw=black] (0.0,0.0) rectangle (0.7,0.7);
\end{tikzpicture}
}

\def\static{
\begin{tikzpicture}[scale=0.25]
\filldraw[fill=white, draw=white] (0.0 ,0.0) rectangle (0.9,0.9);
\filldraw[fill=white, draw=black] (0.1,0.1) rectangle (0.8,0.8);
\end{tikzpicture}
}

\begin{table*}[!t]
\caption{The top 15 most used databases in the reviewed studies; also included are three large databases published in 2016 and 2017.}
\label{tab:databases}
\begin{threeparttable}
\centering
\begin{tabularx}{\textwidth}{
>{\hsize=.23\hsize}X
>{\hsize=.03\hsize}X
>{\hsize=.0075\hsize}X
>{\hsize=.0075\hsize}X
>{\hsize=.0075\hsize}X
>{\hsize=.0075\hsize}X
>{\hsize=.06\hsize}X
>{\hsize=.09\hsize}X
>{\hsize=.14\hsize}X
>{\hsize=.1\hsize}X
>{\hsize=.23\hsize}X
>{\hsize=.04\hsize}X
>{\hsize=.04\hsize}X
}
\toprule
Name & Year & \multicolumn{3}{c}{Modality\tnote{a}} & \multicolumn{3}{c}{Examples\tnote{b}} & \multicolumn{3}{c}{Details on elicitation and annotation} & \multicolumn{2}{c}{Uses\tnote{c}} \\
\cmidrule(lr){3-5} \cmidrule(lr){6-8} \cmidrule(lr){9-11} \cmidrule(lr){12-13}
									   &		  & V 		  & A 		  & P 		  & M 		   & Subjects & Examples & Source & Annotation & Affect (Label) & Target & Transf. \\
\midrule
CK+\cite{lucey2010extended} 			   & 2010 & $\bullet$ &			  & 		  & \sequence & 123  & 593   & Posed (Lab)   & Manual 	 & Categorical (Discrete) & 44 & 6  \\
FER2013\cite{goodfellow2015challenges} & 2013 & $\bullet$ &			  & 		  & \static   & 	 & 35887 & Web search    & Semi-aut. & Categorical (Discrete) & 17 & 23 \\
ImageNet\cite{deng2009imagenet}		   &	 2009 & $\bullet$ & 			  & 		  & \static   &	  	 & 14.2M & Web search    & \multicolumn{2}{c}{(Generic image categories)} & 0 & 27 \\
JAFFE\cite{lyons1998coding}			   & 1998 & $\bullet$ &			  & 		  & \static   & 10   & 219   & Posed (Lab)   & Manual	 & Categorical (Discrete) & 22 & 0  \\
Emo-DB\cite{burkhardt2005database}	   & 2005 &			  & $\bullet$ & 			  & \sequence & 10   & 800   & Posed (Lab)   & Manual	 & Categorical (Discrete) & 17 & 3  \\
VGG-Face\cite{parkhi2015deep}		   & 2015 & $\bullet$ &			  &			  & \static   & 2622 & 2.6M  & Web search	 & \multicolumn{2}{c}{(Generic face images)} & 0 & 18 \\
IEMOCAP\cite{busso2008iemocap}		   & 2008 & $\bullet$ & $\bullet$ &			  & \sequence & 10   & 1039  & Induced (Lab) & Manual	 & Cat./Dim. (Discrete) & 14 & 3  \\
SFEW2\cite{dhall2011static}			   & 2015 & $\bullet$ &			  &			  & \static   & 	 & 1635  & Movies		 & Semi-aut. & Categorical (Discrete) & 9  & 3  \\
DEAP\cite{koelstra2012deap}			   & 2012 & $\bullet$ &			  & $\bullet$ & \sequence & 32   & 40	 & Induced (Lab) & Semi-aut. & Dimensional (Discrete) & 11 & 0  \\
AFEW5\cite{dhall2015video}			   & 2015 & $\bullet$ & $\bullet$ & 			  & \sequence & 	 & 1645  & Movies		 & Semi-aut. & Categorical (Discrete) & 9  & 0  \\
eNTERFACE\cite{martin2006enterface}	   & 2005 & $\bullet$ & $\bullet$ & 			  & \sequence & 42	 & 1166  & Induced (Lab) & Manual	 & Categorical (Discrete) & 8  & 1  \\
AFEW6\cite{dhall2016emotiw}			   & 2016 & $\bullet$ & $\bullet$ & 			  & \sequence & 	 & 1749  & Movies		 & Semi-aut. & Categorical (Discrete) & 8  & 0  \\
RECOLA\cite{ringeval2013introducing}	   & 2013 & $\bullet$ & $\bullet$ & $\bullet$ & \sequence & 23   & 46	 & Spont. (Lab)  & Manual 	 & Dimensional (Cont.) 	  & 8  & 0  \\
CASIA\cite{bao2014building}			   & 2014 & $\bullet$ & $\bullet$ &			  & \sequence & 219  & 2 hr  & TV shows	  	 & Manual	 & Categorical (Discrete) & 7  & 1  \\
SEMAINE\cite{mckeown2012semaine}		   & 2010 & $\bullet$ & $\bullet$ & 			  & \sequence & 20   & 150   & Induced (Lab) & Manual 	 & Dimensional (Cont.) 	  & 7  & 0  \\
\midrule
AffectNet\cite{mollahosseini2017affectnet}&2017& $\bullet$ & 		  &			  & \static   & 450K & 1M	 & Web search	 & Semi-aut. & Cat./Dim. (Discrete)	  & 1  & 0  \\
EmotioNet\cite{benitez2016emotionet}	   & 2016 & $\bullet$ &			  &			  & \static   & 	 & 1M	 & Web search	 & Automatic & Categorical (Discrete) & 1  & 0  \\
AUTOENCODER\cite{gupta2017multi}		   & 2017 & $\bullet$ & $\bullet$ &			  & \sequence & 	 & 6.5M  & Web search	 & \multicolumn{2}{c}{(Non-labeled affective displays)} & 0 & 1 \\
\bottomrule
\end{tabularx}
\begin{tablenotes}
\item[a] Modalities: V = Visual (FER, Body), A = Auditory (SER), P = Physiological (EEG, Peripheral, Other).
\item[b] M = Mode; \static = Static, \sequence = Sequence; Number of subjects given where known.
\item[c] \textit{Target} counts the number of studies that predicted given data; \textit{Transfer} counts the number of studies that used given data for transfer learning. 
\end{tablenotes}
\end{threeparttable}
\end{table*}

Most researchers rely on publicly available databases of affective display as source material for their studies.
Of the {\nDLTotal} studies in our review, only {\prcPrivateDB} involved private databases not available to the public.
A total of {\nDBTotal} different public databases were used across the reviewed literature.
The specifics of these databases have considerable impact on algorithm design for affect recognition, which is why a comprehensive overview of databases and their properties is essential to further understanding of the field.
The main differences lie in the available modalities, the number of subjects and examples, details on how data was acquired, how affect was elicited and annotated, as well as the type of affective states used for labels.
Some databases are more frequently mentioned due to them being featured in competitions.
In Table \ref{tab:databases}, we give a summary of the 15 most commonly used databases in the reviewed studies.

As expected from previous findings, the visual modality is featured most frequently.
Some databases focus exclusively on static FER with discrete labels of categorical affective states.
Here, more recent databases such as FER2013 and SFEW2 tend to contain more examples than older databases such as JAFFE and CK+---this is made possible by resorting to sources like the web and semi-automatic labeling procedures as opposed to manual annotation of data collected in a laboratory.
Audiovisual databases are a second type evident from the literature.
A typical setup for earlier instances (e.g., IEMOCAP, SEMAINE) is a lab-based video recording of subjects, with induced rather than posed affective states.
More recently, physiological sensors have also been included, with various peripheral signals and EEG (e.g., DEAP, RECOLA, and MAHNOB-HCI \cite{soleymani2012multimodal}).
A further approach is to use excerpts from movies and television shows, which can be labeled semi-automatically based on subtitles (e.g., AFEW and CASIA).

Another important aspect of databases is the employed model of affect.
Of the {\nDBTotal} public databases used in the studies covered in our review, {\nCatDatabase} use categorical models ({\prcCatDatabase}), {\nDimDatabase} use dimensional models ({\prcDimDatabase}), and only {\nBothDatabase} use both ({\prcBothDatabase}).
Further, only one database provides unlabeled affective displays (AUTOENCODER).
This heavy reliance of databases on categorical models is also reflected in the models employed in the reviewed studies.
Overall, {\nCatStudies} studies use categorical models ({\prcCatStudies}), {\nDimStudies} studies use dimensional models ({\prcDimStudies}), and only {\nBothStudies} use both ({\prcBothStudies}).
For categorical affective states, every sequence is typically labeled in a discrete fashion with one affective state from a set of pre-defined labels;
whereas, for dimensional affective states, frames are labeled continuously or in discrete steps.
The ambiguity of human affect inherently makes both affect recognition and the labeling process difficult---there is an accuracy limit in the degree of agreement between multiple labelers.
Overall, the trend apparent here goes towards capturing more naturalistic affective displays, as we venture from posed to spontaneous displays.
Also, because of the challenges associated with categorical models, researchers have advocated for further investigating the application of dimensional models in affect recognition and comparing them with categorical models \cite{calvo2010affect}, \cite{gunes2013categorical}, \cite{bugnon2017dimensional}.
Unfortunately, at this stage, the number of examples per dataset does not see a clear upward trend yet and only few deep learning studies covered in our review investigate both types of models. 

Unlabeled databases are used exclusively for transfer learning.
When considering large general-purpose databases like ImageNet, the idea is to learn general low-level descriptors that help to improve initial model parameters.
For FER, more relevant databases of unlabeled face images (e.g., VGG-Face) can be used.
Smaller, labeled databases such as FER2013 are also used frequently for supervised pre-training.
Note that these data sources for transfer learning are primarily focused on static examples of the visual modality.

Databases published in 2016 and 2017 aim to provide sufficient training data for deep learning models.
Compared to older databases, they contain many more examples, which is made possible by (semi-)automating the labeling process, or providing unlabeled examples.
Three such databases are included in Table \ref{tab:databases}.
Both AffectNet and EmotioNet are large web-based databases, each at around 1M labeled images.
Notably, AffectNet includes both dimensional and categorical labels, encouraging studies to bridge the gap between affect representations.
The AUTOENCODER dataset is the largest face video dataset with 6.5M examples; the dataset contains only 2777 labeled examples and is thus largely unlabeled.
It can serve the purpose of unsupervised pre-training or semi-supervised learning \cite{gupta2017multi}.

\begin{table*}[!h]
\caption{Winning entries at affect recognition competitions (2013-2017).}
\label{tab:competitions}
\begin{threeparttable}
\centering
\begin{tabularx}{\textwidth}{
>{\hsize=.18\hsize}X
>{\hsize=.18\hsize}X
>{\hsize=.07\hsize}X
>{\hsize=.20\hsize}X
>{\hsize=.04\hsize}X
>{\hsize=.04\hsize}X
>{\hsize=.04\hsize}X
>{\hsize=.30\hsize}X
}
\toprule
\multicolumn{2}{c}{Competition} & \multirow{2}{*}{Database} & \multirow{2}{*}{Winner} & \multicolumn{3}{c}{Deep feature learning} & \multirow{2}{*}{Evaluation statistic\tnote{b}} \\
\cmidrule(lr){1-2} \cmidrule(lr){5-7}
Name & Sub-Challenge\tnote{a} & & & Spatial & Temp. & Joint & \\
\midrule	
AVEC 2013	& Fully cont. A/V	& AVID		& Meng et al. \cite{meng2013depression}		&			&			&			& CC: 0.141							\\
EmotiW 2013	& A/V				& AFEW3		& Kahou et al. \cite{kahou2013combining}	& $\bullet$	&			&			& Accuracy: 41.0\%					\\
ICML 2013	& Static FER			& FER2013	& Tang et al. \cite{tang2013deep}			& $\bullet$	&			&			& Accuracy: 71.2\%					\\
INTERSPEECH'13 & SER			& GEMEP		& Gosztolya et al. \cite{gosztolya2013detecting}& 		& 			& 			& Accuracy: 73.5\% (A), 63.3\% (V)	\\
AVEC 2014	& Fully cont. A/V	& AVID		& Kachele et al.	 \cite{kachele2014inferring}&			&			&			& CC: 0.63 (A), 0.58 (V), 0.57 (D)	\\	
EmotiW 2014	& A/V				& AFEW4		& Liu et al.	 \cite{liu2014combining}		& $\bullet$	&			&			& Accuracy: 50.4\%					\\
AVEC 2015	& Fully cont. A/V/P	& RECOLA 	& He et al. \cite{he2015multimodal} 		&   		&  			& $\bullet$	& CCC: 0.747 (A), 0.609 (V)			\\
EmotiW 2015	& A/V				& AFEW5		& Yao et al.	 \cite{yao2015capturing}		& $\bullet$	&			&			& Accuracy: 53.8\%					\\
EmotiW 2015	& Static FER			& SFEW2		& Kim et al.	 \cite{kim2015hierarchical}		& $\bullet$	&			&			& Accuracy: 61.6\%					\\
AVEC 2016	& Fully cont. A/V/P	& RECOLA 	& Brady et al. \cite{brady2016multi} 		& $\bullet$	& $\bullet$	&			& CCC: 0.77 (A), 0.687 (V)			\\
EmotiW 2016	& A/V				& AFEW6		& Fan et al. \cite{fan2016video}				& $\bullet$	& $\bullet$	&			& Accuracy: 59.0\%					\\
EmotiW 2016	& Group-level FER	& HAPPEI		& Li et al.	\cite{li2016happiness}			& $\bullet$	&			& 			& RMSE: 0.82						\\
AVEC 2017	& Fully cont. A/V	& SEWA		& Chen et al. \cite{chen2017multimodal}		& $\bullet$	& $\bullet$ & $\bullet$	& CCC: 0.68 (A), 0.76 (V), 0.51 (L)	\\
EmotiW 2017	& A/V				& AFEW7		& Hu et al. \cite{hu2017learning}			& $\bullet$	&			&			& Accuracy: 60.3\%					\\
EmotiW 2017	& Group-level FER	& GAF		& Tan et al. \cite{tan2017group}				& $\bullet$	&			&			& Accuracy: 80.9\%					\\
\bottomrule
\end{tabularx}
\begin{tablenotes}
\item[a] Modalities: V = Visual (FER, Body), A = Auditory (SER), P = Physiological (EEG, Peripheral, Other).
\item[b] Reporting statistics used in competitions: CC = Correlation coefficient, CCC = Concordance correlation coefficient, A = Arousal, V = Valence, D = Dominance, L = Likability, RMSE = Root mean square error.
\end{tablenotes}
\end{threeparttable}
\end{table*}

In this review, we have thus far avoided directly comparing studies based on their recognition accuracies.
Our reasoning is that even if both studies focus on the same dataset, differing test and training sets as well as evaluation statistics lead to difficulties in fairly comparing results.
This situation is different for organized competitions, where the criteria are clearly defined and results independently verified.
To illustrate state-of-the-art recognition accuracies, Table \ref{tab:competitions} reports the winning entries at affect recognition competitions from 2013 to 2017.
These can generally be thought of as the latest and best results for the respective datasets at publication time.
We employ the classification scheme introduced in Table \ref{tab:literature} and used throughout this review to indicate the use of deep learning in these studies.

It is evident that since 2015, all winning entries at these competitions have made use of deep learning.
Evaluation statistics are given by recognition accuracy for discrete affect recognition (EmotiW), and correlation coefficients for continuous affect recognition (AVEC).
Progress in recognition accuracies is measurable:
On the AFEW dataset, which is extended with additional data every year, recognition accuracies have increased from 41\% in 2013 to 60.3\% in 2017.

\section{Discussion}

In this review, we have seen that DNNs are part of most state-of-the-art affect recognition systems. 
They are applied for learning of (i) spatial feature representations, (ii) temporal feature representations, and (iii) joint feature representations for multimodal data.
Considering the recent trend towards continuous and multimodal prediction of spontaneous affective displays in the wild, deep learning is generally well suited to address the challenges faced by such systems.
Particularly, our results show that out of the {\nComparison} studies reporting comparisons between shallow and deep architectures, {\prcDeepBetter} reported that deep learning can lead to improvements over conventional approaches.

However, in comparison to related fields like object detection and ASR, the impact of deep learning has not yet been fully felt.
This can in part be attributed to the higher difficulty and inherent ambiguity of affective displays---although many databases now provide annotations from multiple labelers, the ground truth tends to be unreliable and datasets are often imbalanced.
A major obstacle is the relatively small size of labeled datasets.
It hinders the ability of deep models to generalize well, and makes it difficult to train large models.
In this vein, regularization techniques from general deep learning research \cite{kukavcka2017regularization} are one future research avenue.
However, as of this writing, employed techniques such as knowledge transfer from related disciplines and data augmentation ultimately cannot fully make up for the lack of data.

Although we notice a trend towards larger labeled datasets, new approaches are needed to deal with this issue.
The labeling process for video-based affective datasets, especially with continuous annotations, is expensive and cannot easily be automated.
Unsupervised and semi-supervised learning are promising trends in this regard.
While unsupervised learning allows models to learn better initial parameters from unlabeled datasets, fine-tuning with labeled data is still required to "guide" the model towards its learning goal.
The idea of semi-supervised learning is to label only a fraction of the examples, avoiding the expensive labeling process for the most part.
These labeled examples can then be leveraged to learn from the remaining, much larger unlabeled part of the database that was acquired at a lower cost.
Gupta et al. \cite{gupta2017multi} demonstrated the feasibility of semi-supervised learning for affective computing, achieving promising results using 2777 labeled face videos to learn from the AUTOENCODER dataset of 6.5M unlabeled face video clips.
In 2018, supervised pre-training was successfully used at Facebook \cite{mahajan2018exploring} to transfer knowledge of 3.5B public Instagram images and hashtags for generic image classification and object detection.
This development suggests that affect recognition too may benefit from much larger datasets. 
Unsupervised learning has also proved successful in SER--for example, Deng et al. \cite{deng2017universum} were able to improve the training process on labeled data by exploiting knowledge from unlabeled data using AEs.  

In most state-of-the-art affect recognition models, handcrafted features still play an important role.
While they are typically outperformed by DNN-learned features in direct comparison, some challenge-winning models rely on hybrid architectures that can take advantage of complementarities between the two \cite{yan2016multi}, \cite{brady2016multi}.
This can be attributed to multiple factors:
Handcrafted features are readily available, widely recognized, and well-designed for specific applications such as FER and SER.
On the other hand, deep models have not yet been established as methods of affect recognition.
There are few specialized architectures researchers can draw from, which is why many fall back to ones from object recognition.
Furthermore, the number of examples to draw from is not large enough to learn truly expressive deep features.
Research suggests that the size of a dataset can be a bottleneck for performance of deep learning models \cite{sun2017revisiting}.
As these circumstances change, larger and more expressive deep models specialized for affect recognition will have a chance of being established.
Pre-trained deep models for FER and SER could become readily available as feature extractors, similar to handcrafted features.

As we have seen, affect recognition models are comprised of multiple components, including the steps of spatial and temporal feature learning as well as an additional fusion mechanism in the case of multimodal systems.
We find that as of now, DNNs are generally applied in an isolated way to manage individual components---hence, a combination of multiple DNN-based components comes to mind.
In fact, one ideal of classic deep learning research is the idea of integrating multiple components into globally trainable systems \cite{lecun1998gradient}.
Some recent contributions have shown that affect recognition systems can combine multiple DNN-based components.
As discussed throughout this review, the combination of CNN for feature learning and LSTM to learn the temporal context is most widely used ({\nCNNLSTM} studies).
However, most of these approaches train individual components separately (e.g., \cite{kahou2015recurrent}, \cite{khorrami2016deep}), which may lead to suboptimal performance considering the whole system.
Future multimodal affect recognition systems are likely to be trained in an end-to-end fashion, which aligns training of the entire network closer with the true performance measure \cite{graves2014towards} but requires a larger amount of training examples.
The feasibility and effectiveness of global end-to-end training has been demonstrated in two recent studies:
Trigeorgis et al. \cite{trigeorgis2016adieu} combined CNN and LSTM to learn features and temporal context directly from speech signals.
Similarly, Tzirakis et al. \cite{tzirakis2017end} used the CNN-LSTM architecture to learn features and temporal context directly from raw audio-visual signals.

Another recent trend from deep learning research that lends itself to affect recognition of sequential data is the use of attention mechanisms \cite{vaswani2017attention}.
They allow models to learn to pay more attention to promising segments of, say, a video of affective display.
Mirsamadi et al. \cite{mirsamadi2017automatic} found that even a simple attention mechanism contributes to SER by, for example, learning to ignore silent frames---an improvement that is easily interpretable and adds 1-2\% to model accuracy.

\begin{figure}[!t]
\centering
\includegraphics[width=3.2in]{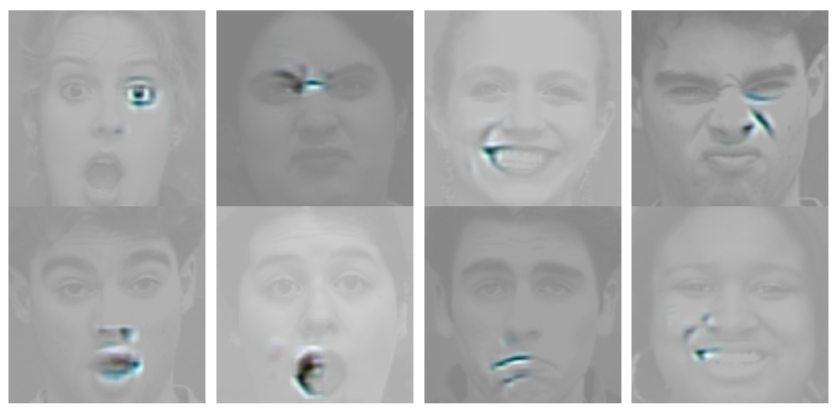}
\caption{CNNs learn high-level features similar to Action Units in FER (taken from \cite{breuer2017deep} with permission).}
\label{fig:au}
\end{figure}

In general, DNNs are often described as "black boxes", as their inner workings are complex and difficult to interpret.
Recent research has attempted to improve this situation, for example by offering ways to visualize the activations of individual filters in CNNs \cite{zeiler2014visualizing}.
Multiple studies have followed this approach to understand how CNNs recognize facial expressions, finding that higher layers learn concepts similar to Action Units (e.g., \cite{khorrami2015deep}, \cite{breuer2017deep}), validating previous research done in FER---Fig. \ref{fig:au} features some examples.
Similarly, for SER, Tzirakis et al. \cite{tzirakis2017end} found that LSTM cells learn representations similar to well-known prosodic features.
Such results indicate that the study of DNNs in affective computing could contribute to interdisciplinary emotion research by giving a further perspective on affective displays and their representation in general.

The question of categorical versus dimensional representations of affective states remains unsolved, as no deciding trend emerges from the literature.
In practice, specific details of available modalities (e.g., static images or video data) and requirements regarding predictions (e.g., continuous or video-level) are relevant factors, and available labels on common databases widely dictate which representation is used.
As can be seen in our review, the majority of studies still rely on categorical models of affect (\prcCatStudies), particularly in spatial feature learning (\prcCatSpatial).
However, given the nature of sensor data in typical HCI scenarios, it can be expected that dimensional representations, and combinations of categorical and dimensional models of affect will become even more relevant as a more natural way of dealing with continuous data \cite{gunes2013categorical}.
This already becomes apparent in joint feature learning and to some extent in temporal feature learning, where dimensional models are already employed in {\prcDimJoint} and {\prcDimTemporal} of the reviewed studies, respectively.
Databases such as AffectNet \cite{mollahosseini2017affectnet}, which provide categorical as well as dimensional annotations of affect in the wild, are an important step for furthering the investigation of dimensional models and for better integrating the context of affective display into human affect recognition.

\section*{Acknowledgments}

This research was supported by an Australian Government Research Training Program (RTP) Scholarship.

\ifCLASSOPTIONcaptionsoff
  \newpage
\fi



\bibliographystyle{assets/IEEEtran}
\bibliography{paper}

%

\begin{IEEEbiography}[{\includegraphics[width=1in,height=1.25in,clip,keepaspectratio]{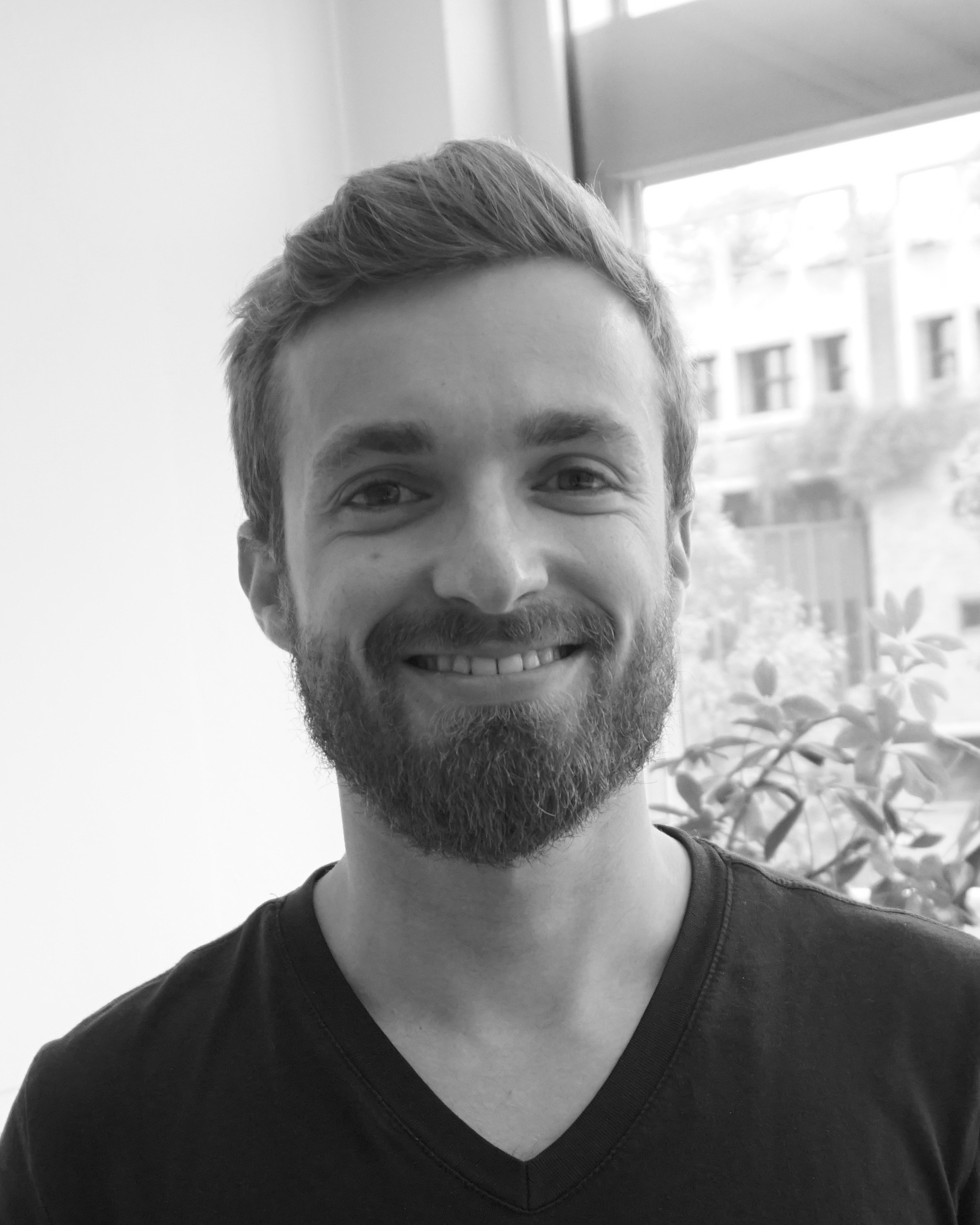}}]{Philipp V. Rouast}
received the B.Sc. and M.Sc. degrees in Industrial Engineering from Karlsruhe Institute of Technology, Germany, in 2013 and 2016 respectively.
He is currently working towards the Ph.D. degree in Information Systems and is a graduate research assistant at The University of Newcastle, Australia.
His research interests include deep learning, affective computing, HCI, and related applications of computer vision.
Find him at \url{https://prouast.github.io}.
\end{IEEEbiography}

\begin{IEEEbiography}[{\includegraphics[width=1in,height=1.25in,clip,keepaspectratio]{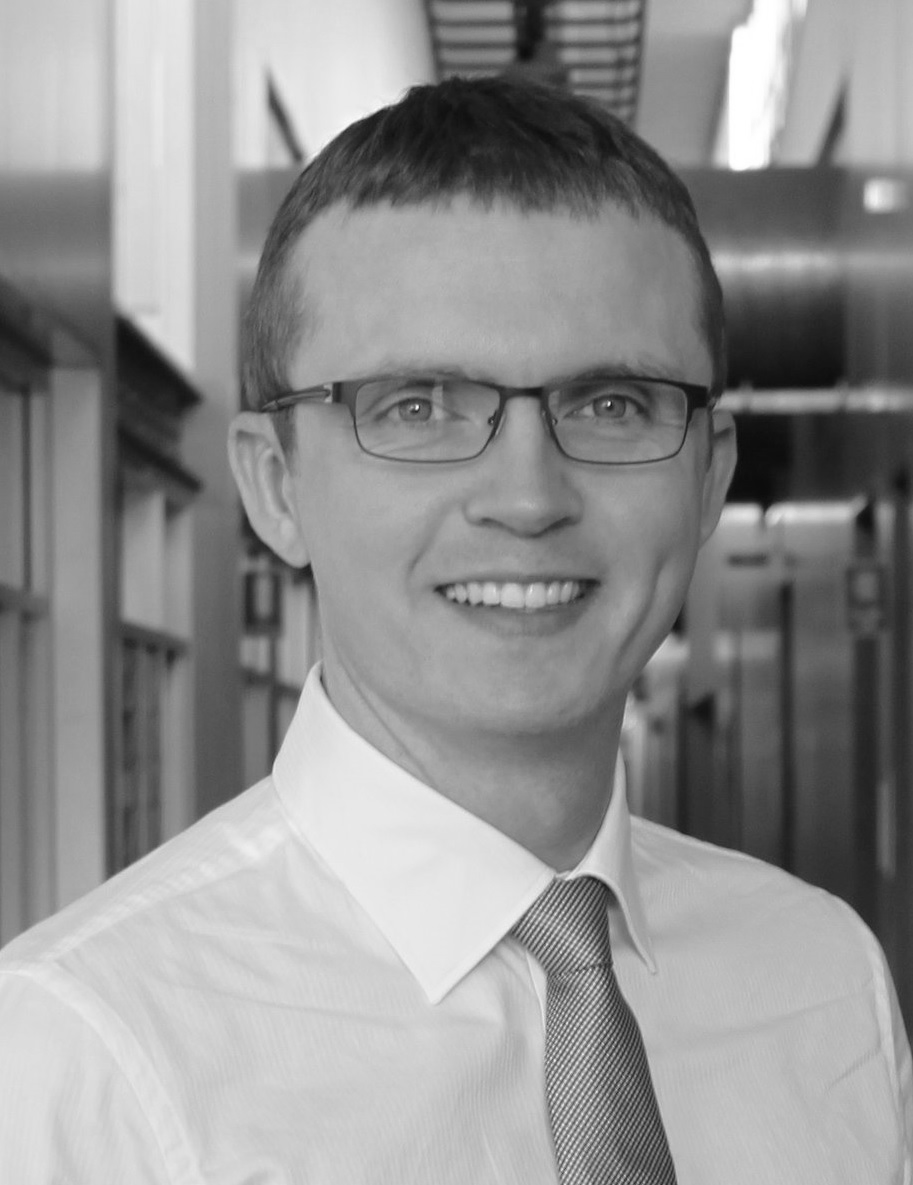}}]{Marc T. P. Adam}
is a Senior Lecturer in Computing and Information Technology at the University of Newcastle, Australia.
In his research, he investigates the interplay of human users' cognition and affect in human-computer interaction.
He received an undergraduate degree in Computer Science from the University of Applied Sciences W{\"u}rzburg, Germany, and a PhD in Economics of Information Systems from Karlsruhe Institute of Technology, Germany.
\end{IEEEbiography}

\begin{IEEEbiography}[{\includegraphics[width=1in,height=1.25in,clip,keepaspectratio]{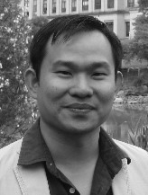}}]{Raymond Chiong}
(M'05-SM'14) is a Senior Lecturer at the University of Newcastle, Australia.
He received his MSc from the University of Birmingham, UK, and PhD from the University of Melbourne, Australia.
His research interests include evolutionary game theory, optimization, data analytics, and modeling of complex adaptive systems.
He is the Editor-in-Chief of the Journal of Systems and Information Technology, an Editor of Engineering Applications of Artificial Intelligence, and an Associate Editor of the Computational Intelligence Magazine.
\end{IEEEbiography}




\end{document}